\documentclass[final]{cvpr}

\usepackage{times}
\usepackage{epsfig}
\usepackage{graphicx}
\usepackage{amsmath}
\usepackage{amssymb}

\usepackage{amsmath,amsfonts,bm}

\def\eqref#1{equation~\ref{#1}}

\def\1{\bm{1}}

\newcommand{\test}{\mathcal{D_{\mathrm{test}}}}

\DeclareMathAlphabet{\mathsfit}{\encodingdefault}{\sfdefault}{m}{sl}
\SetMathAlphabet{\mathsfit}{bold}{\encodingdefault}{\sfdefault}{bx}{n}

\DeclareMathOperator*{\argmin}{arg\,min}

\usepackage[utf8]{inputenc} 
\usepackage[T1]{fontenc}    
\usepackage{url}            
\usepackage{booktabs}       
\usepackage{enumitem}
\usepackage{amsfonts}       
\usepackage{nicefrac}       
\usepackage{microtype}      
\usepackage{times}
\usepackage{epsfig}
\usepackage{graphicx}
\usepackage{comment}
\usepackage{multirow}
\usepackage{multicol}
\usepackage{pifont}
\usepackage{soul,color}
\usepackage{subcaption}
\usepackage[sort&compress,square,numbers,comma]{natbib}

\usepackage{enumitem}
\setitemize{noitemsep,topsep=0pt,parsep=0pt,partopsep=0pt}
\setlist{noitemsep,topsep=0pt,parsep=0pt,partopsep=0pt}

\newcommand{\change}[1]{{#1}}

\newcommand{\cmark}{\ding{51}}%
\usepackage[pagebackref=true,breaklinks=true,colorlinks,bookmarks=false]{hyperref}

\def\cvprPaperID{7828} 
\def\confYear{CVPR 2021}

\begin{document}


\title{StEP: Style-based Encoder Pre-training for Multi-modal Image Synthesis}

\author{
  Moustafa Meshry\quad 
    Yixuan Ren\quad 
    Larry S. Davis\quad
    Abhinav Shrivastava\\[0.3em]
    University of Maryland, College Park\mbox{ }
}

\maketitle

\begin{abstract}

We propose a novel approach for multi-modal Image-to-image (I2I) translation.
To tackle the one-to-many relationship between input and output domains, previous works use complex training objectives to learn a latent embedding, jointly with the generator, that models the variability of the output domain.
In contrast, we directly model the style variability of images, independent of the image synthesis task.
Specifically, we pre-train a generic style encoder using a novel proxy task to learn an embedding of images, from arbitrary domains, into a low-dimensional style latent space.
The learned latent space introduces several advantages over previous traditional approaches to multi-modal I2I translation.
First, it is not dependent on the target dataset, and generalizes well across multiple domains.
Second, it learns a more powerful and expressive latent space, which improves the fidelity of style capture and transfer.
The proposed style pre-training also simplifies the training objective and speeds up the training significantly.
Furthermore, we provide a detailed study of the contribution of different loss terms to the task of multi-modal I2I translation, and propose a simple alternative to VAEs to enable sampling from unconstrained latent spaces.
Finally, we achieve state-of-the-art results on six challenging benchmarks with a simple training objective that includes only a GAN loss and a reconstruction loss.

\end{abstract}

\vspace{-0.17in}
\section{Introduction}
\vspace{-0.07in}

Image-to-Image (I2I) translation is the task of transforming images from one domain to another (e.g., semantic maps $\rightarrow$ scenes, sketches $\rightarrow$ photo-realistic images, etc.).
Many problems in computer vision and graphics can be cast as I2I translation, such as
photo-realistic image synthesis~\cite{chen2017photographic,isola2017image,wang2017high},
super-resolution~\cite{ledig2017photo},
colorization~\cite{zhang2016colorful,zhang2017real},
and inpainting~\cite{pathak2016context}.
Therefore, I2I translation has recently received significant attention in the literature.
%
One main challenge in I2I translation is the multi-modal nature for many such tasks --
the relation between an input domain $A$ and an output domain $B$ is often times one-to-many, where a single input image $I_i^A \in A$ can be mapped to different output images from domain $B$.
For example, a sketch of a shoe or a handbag can be mapped to corresponding objects with different colors or styles, or a semantic map of a scene can be mapped to many scenes with different appearance, lighting and/or weather conditions.
%
Since I2I translation networks typically learn one-to-one mappings due to their deterministic nature, an extra input is required to specify an output mode to which an input image will be translated.
Simply injecting extra random noise as input proved to be ineffective as shown in~\cite{isola2017image,zhu2017toward}, where the generator network just learns to ignore the extra noise and collapses to a single or few modes (which is one form of the mode collapse problem).
To overcome this problem, Zhu \etal\cite{zhu2017toward} proposed \emph{BicycleGAN}, which trains an encoder network $E$, jointly with the I2I translation network, to encode the distribution of different possible outputs into a latent vector $z$, and then learns a deterministic mapping $G: (A, z) \rightarrow B$.
So, depending on the latent vector $z$, a single input $I^A_i \in A$ can be mapped to multiple outputs in $B$.
%
While BicycleGAN requires paired training data, several works
, like MUNIT~\cite{huang2018multimodal} and DRIT~\cite{lee2018diverse},
extended it to the unsupervised case, where images in domains $A$ and $B$ are not in correspondence (`unpaired').
One main component of unpaired I2I is a cross-cycle consistency constraint, 
where the network generates an intermediate output by swapping the styles of a pair of images, then swaps the style between the intermediate output again to reconstruct the original images. This enforces that the latent vector $z$ preserves the encoded style information when translated from an image $i$ to another image $j$ and back to image $i$ again.
This constraint can also be applied to paired training data, where it encourages style/attribute transfer between images.
However, training BicycleGAN~\cite{zhu2017toward} or its unsupervised counterparts~\cite{huang2018multimodal,lee2018diverse} is not trivial.
For example, BicycleGAN combines the objectives of both conditional Variational Auto-Encoders (cVAEs)~\cite{sohn2015learning} and a conditional version of Latent Regressor GANs (cLR-GANs)~\cite{donahue2016adversarial,dumoulin2016adversarially}  to train their network.
The training objective of~\cite{huang2018multimodal,lee2018diverse} is even more involved to handle the unsupervised setup.

In this work, we propose a novel weakly-supervised pre-training strategy to learn an expressive latent space for the task of multi-modal I2I translation.
While end-to-end training of the encoder network $E$ with the I2I translation network poses a convenience, we show that it can be advantageous to break down the training into proxy tasks.
In specific, we show both quantitatively and qualitatively that the proposed pre-training yields the following \textbf{advantages}:
\begin{itemize}[leftmargin=*,noitemsep,topsep=0pt,parsep=0pt]
  \item It learns a more powerful and expressive latent space. Specifically, we show that: (1) Our pre-trained latent space captures uncommon styles that are not well represented in the training set, while baselines like BicycleGAN~\cite{zhu2017toward} and MUNIT~\cite{huang2018multimodal} fail to do so and instead tend to simplify such styles/appearances to the nearest common style in the train set.
      (2) Pre-training yields more faithful style capture and transfer.
      (3) Finally, the better expressiveness of the pre-trained latent space leads to more complex style interpolations compared to the baselines.
  \item The learned style embedding is not dependent on the target dataset and generalizes well across many domains, which can be useful especially when having limited training data.
  \item Style pre-training simplifies the training objective by requiring fewer losses, which also speeds up the training.
  \item Our approach improves the training stability and the overall output quality and diversity.
\end{itemize}
We note that our proposed style pre-training is weakly-supervised and doesn't require any manual labeling. Instead, it relies on a pre-trained VGG network~\cite{johnson2016perceptual} to provide training supervision.
Our approach is inspired by and extends the work of Meshry~\etal\cite{meshry2019neural} which utilizes a staged training strategy to re-render scenes under different lighting, time of day, and weather conditions.
Our work is also inspired by the standard training paradigm in visual recognition of first pre-training on a proxy task, either large supervised datasets (e.g., ImageNet)~\cite{krizhevsky2012imagenet,mahajan2018exploring,sun2017revisiting} or unsupervised tasks~(e.g.,~\cite{doersch2015unsupervised,noroozi2016unsupervised}), and then finetuning (transfer learning) on the desired task. Similarly, we propose to pre-train the encoder using a proxy task that encourages capturing style into a latent space. Our goal is to highlight the benefits of encoder pre-training and demonstrate its effectiveness for multi-modal image synthesis. 
In particular, we make the following \textbf{contributions}:
%
\begin{itemize}[leftmargin=*,noitemsep,topsep=0pt,parsep=0pt]
    \item We propose to pre-train an encoder to learn a low-dimensional projection of Gram matrices (from Neural Style Transfer) and show that the pre-trained embedding is effective for multi-modal I2I translation, and that it simplifies and stabilizes the training.
    \item We show that the pre-trained latent embedding is not dependent on the target domain and generalizes well to other domains (transfer learning).
    \item We provide a study of the importance of different loss terms for multi-modal I2I translation network.
    \item We propose an alternative to enable sampling from a latent space instead of enforcing a prior as done in VAE training.
    \item We achieve state-of-the art results on six benchmarks in terms of \change{style capture and transfer, and diversity of results}. 
\end{itemize}

\vspace{-0.20in}
\section{Related work}
\vspace{-0.07in}

\noindent
\textbf{Deep generative models.}
There has been incredible progress in the field of image synthesis using deep neural networks.
In its unconditional setting, a decoder network learns to map random values drawn from a prior distribution (typically Gaussian) to output images.
Variational Auto-Encoders (VAEs)~\cite{kingma2013auto} assume a bijection mapping between output images and some latent distribution and learn to map the latent distribution to a unit Gaussian using the reparameterization trick.
Alternatively, Generative Adversarial Networks (GANs)~\cite{goodfellow2014generative} directly map random values sampled from a unit Gaussian to images, while using a discriminator network to enforce that the distribution of generated images resembles that of real images. Recent works proposed improvements to stabilize the training~\cite{gulrajani2017improved,karnewar2019msg,mao2017least,radford2015unsupervised} and improve the quality and diversity of the output~\cite{karras2017progressive,karras2018style}.
Other works combine both VAEs and GANs into a hybrid VAE-GAN model~\cite{larsen2015autoencoding,rosca2017variational}.

\noindent
\textbf{Conditional image synthesis.}
Instead of generating images from input noise, the generator can be augmented with side information in the form of extra conditional inputs. 
For example, Sohn \etal \cite{sohn2015learning} extended VAEs to their conditional setup (cVAEs).
Also, GANs can be conditioned on different information, like class labels~\cite{mirza2014conditional,odena2017conditional,van2016conditional},
language description~\cite{mansimov2015generating,reed2016generative}, or an image from another domain~\cite{chen2017photographic,isola2017image}. The latter is called Image-to-Image translation.

\noindent
\textbf{Image-to-Image (I2I) translation.}
I2I translation is the task of transforming an image from one domain, such as a sketch, into another domain, such as photo-realistic images.
While there are regression-based approaches to this problem~\cite{chen2017photographic,hoshen2018nam}, significant successes in this field are based on GANs and the influential work of pix2pix~\cite{isola2017image}.
Following the success of pix2pix~\cite{isola2017image}, I2I translation has since been utilized in a large number of tasks, like inpainting~\cite{pathak2016context}, colorization~\cite{zhang2016colorful,zhang2017real}, super-resolution~\cite{ledig2017photo}, rendering~\cite{martinbrualla2018lookingood,meshry2019neural,thies2019deferred}, and many more~\cite{dong2017semantic,wang2016generative,zhang2017age}.
There has also been works to extend this task to the unsupervised setting~\cite{hoshen2018nam,kim2017learning,liu2017unsupervised,ma2018exemplar,royer2017xgan,zhu2017unpaired}, to multiple domains~\cite{choi2018stargan,choi2020stargan}, and to videos~\cite{chan2018everybody,wang2018video}.

\noindent
\textbf{Multi-modal I2I translation.}
Image translation networks are typically deterministic function approximators that learn a one-to-one mapping between inputs and outputs. To extend I2I translation to the case of diverse multi-modal outputs, 
Zhu \etal\cite{zhu2017toward} proposed the BicycleGAN framework that learns a latent distribution that encodes the variability of the output domain and conditions the generator on this extra latent vector for multi-modal image synthesis.
Wang \etal\cite{wang2017high,wang2018video} learn instance-wise latent features for different objects in a target image, which are clustered after training to find $f$ fixed modes for different semantic classes. At test time, they sample one of the feature clusters for each object to achieve multi-modal synthesis.
Other works extended the multi-modal I2I framework to the unpaired setup, where images from the input and output domains are not in correspondence~\cite{almahairi2018augmented,huang2018multimodal,lee2018diverse}, by augmenting BicycleGAN with different forms of a cross-cycle consistency constraint between two unpaired image pairs. 
In our work, we focus on the supervised setting of multi-modal I2I translation. We propose a pre-training strategy to learn a latent distribution that encodes the variability of the output domain. The learned distribution can be easily adapted to new unseen datasets with simple finetuning, instead of training from scratch. 


\section{Approach}
\label{sec:approach}
%
%
Current multi-modal image translation networks require an extra input $z$ that allows for modelling the one-to-many relation between an input domain $A$ and an output domain $B$ as a one-to-one relation from a pair of inputs $(A, z) \rightarrow B$.
In previous approaches, there has been a trade-off between simplicity and effectiveness for providing the input $z$.
On one hand, providing random noise as the extra input $z$ maintains a simple training objective (same as in pix2pix~\cite{isola2017image}). However,~\cite{isola2017image,zhu2017toward} showed that the generator has little incentive to utilize the input vector $z$ since it only encodes random information, and therefore the generator ends up ignoring $z$ and collapsing to one or few modes.
On the other hand, \emph{BicycleGAN}~\cite{zhu2017toward} combines the objectives of both conditional Variational Auto-Encoder GANs (cVAE-GAN) and conditional Latent Regressor GANs (cLR-GAN) to learn a latent embedding $z$ simultaneously with the generator $G$.
Their training enforces two cycle consistencies: $B \rightarrow z \rightarrow \hat{B}$  and $z \rightarrow \tilde{B} \rightarrow \hat{z}$.
This proved to be very effective, but the training objective is more involved, which makes the training slower.
Also, since the latent embedding is being trained simultaneously with the the generator, hyper-parameter tuning becomes more critical and sensitive.
The training objective of more recent works (e.g., \cite{huang2018multimodal,lee2018diverse}) is even more complicated.
We aim to combine the best of both worlds: an effective training of a latent embedding that models the distribution of possible outputs, while retaining a simple training objective. 
This would allow for faster and more efficient training, as well as less sensitivity to hyper-parameters.
We observe that the variability in many target domains can be represented by the style diversity of images in the target domain $B$, where the style is defined in terms of the Gram matrices used in the Neural Style Transfer literature~\cite{gatys2016image}.
However, using Gram matrices directly to represent styles is not feasible due to its very high dimensionality.
So, instead we learn an embedding by separately training an encoder network $E$ on an auxiliary task to optimize for $z = E(I^B)$ capturing the style of an image $I^B$.
Visualizing the pre-trained latent space shows that our pre-trained encoder models different modes of the output distribution (e.g., different colors, lighting and weather conditions, \dots etc.) as clusters of images with similar styles as shown in $\S$\ref{sec:pretrained_latent_vis}.
Then, to synthesize an image $\hat{I}^B = G(I^A, z)$, the input latent can be used to clearly distinguish the style cluster to which the output belongs.
This makes for an effective and more stable training of the generator $G$, since $G$ is just required to discover the correlation between output images and their corresponding style embedding $z$.
Moreover, experimental evaluation shows that the proposed style-based pre-training yields better results in terms of more faithful style capture and transfer, as well as better output quality and diversity.

\begin{figure}[t!]
    \centering
        \includegraphics[width=.94\linewidth]{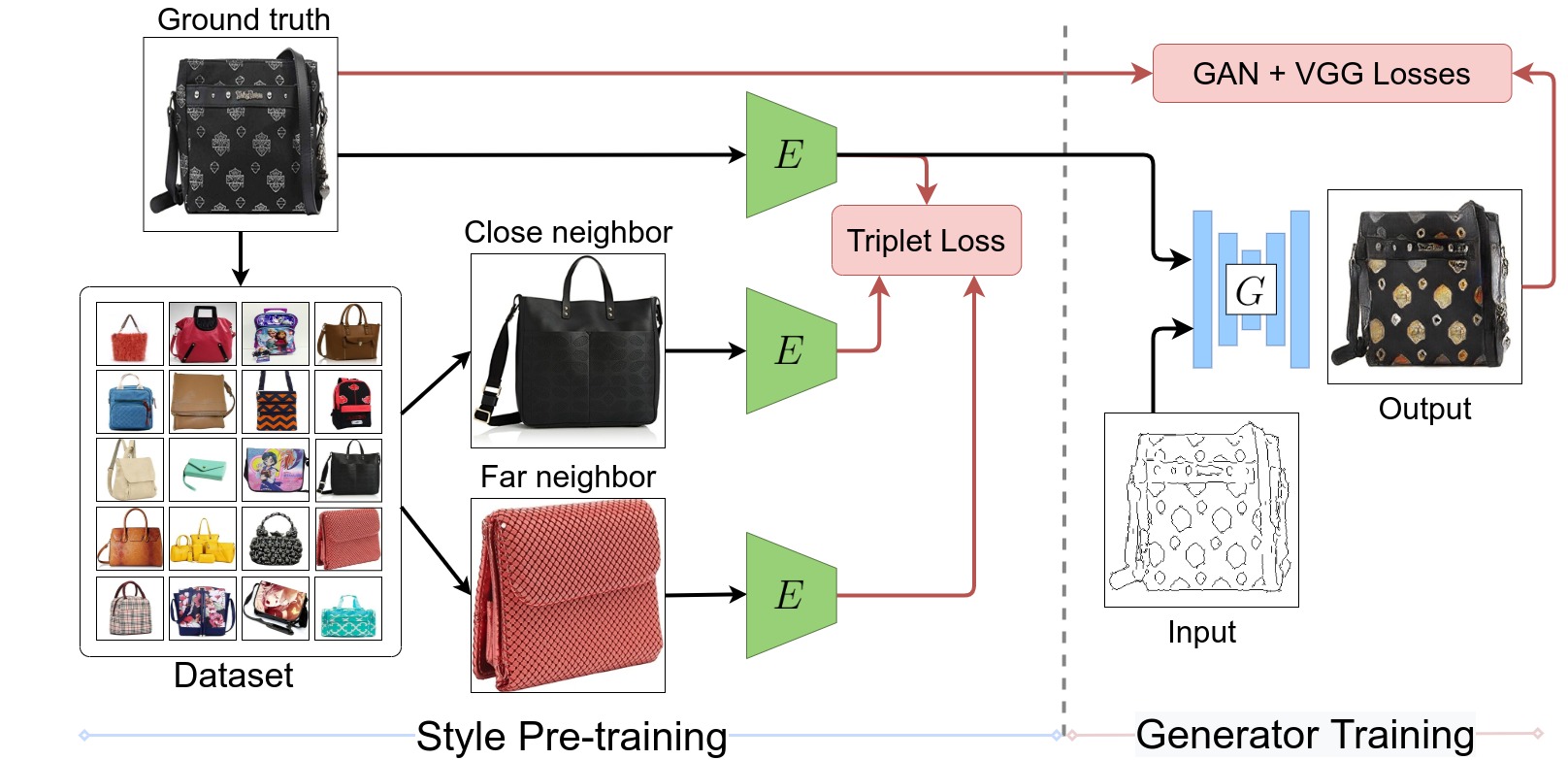}
    \caption{Overview of our training pipeline. Stage 1: pre-training the style encoder $E$ using a triplet loss. Stages 2, 3: training the generator $G$, and finetuning both $G, E$ together using \emph{GAN} and reconstruction losses.}
    \label{fig:training}
    \vspace{-0.24in}
\end{figure}

To incorporate this into BicycleGAN~\cite{zhu2017toward}, we replace the simultaneous training of the encoder $E$ and the generator $G$ with a staged training (Figure~\ref{fig:training}) as follows:
\begin{itemize}[leftmargin=*]
    \item \textbf{Stage 1}: Pre-train $E$ on a proxy task that optimizes an embedding of images in the output domain $B$ into a low-dimensional style latent space, such that images with similar styles lie closely in that space (i.e., clustered).
    \item \textbf{Stage 2}: Train the generator network $G$ while fixing the encoder $E$, so that $G$ learns to associate the style of output images to their deterministic style embedding $z = E(I^B)$.
    \item \textbf{Stage 3}: Finetune both the $E$ and $G$ networks together, allowing for the style embedding to be further adapted to best suit the image synthesis task for the target domain.
\end{itemize}
%
Next, we explain how to pre-train the style encoder network $E$ in $\S$\ref{sec:pretraining}, and how to train the generator $G$ using the pre-learned embeddings ($\S$\ref{sec:g_training}).
Finally, we demonstrate the generalization of pre-training the style encoder $E$ in $\S$\ref{sec:e_generalization}.



\begin{figure*}[t!]
    \centering
    \includegraphics[width=0.67\linewidth]{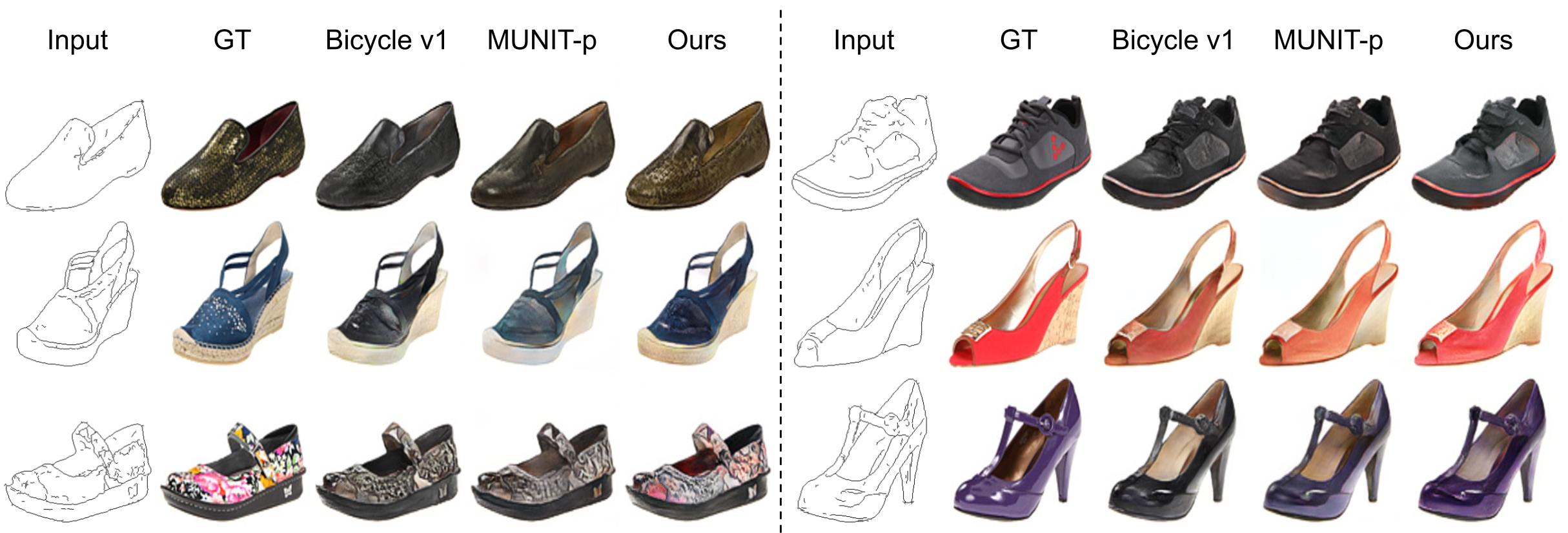}
    \vspace{-0.2cm}
    \caption{Qualitative comparison with baselines. Ours better matches the ground truth (GT) style.}
    \label{fig:qualitative}
    \vspace{-0.2in}
\end{figure*}

\begin{figure*}[t!]
    \centering
    \includegraphics[width=0.68\linewidth]{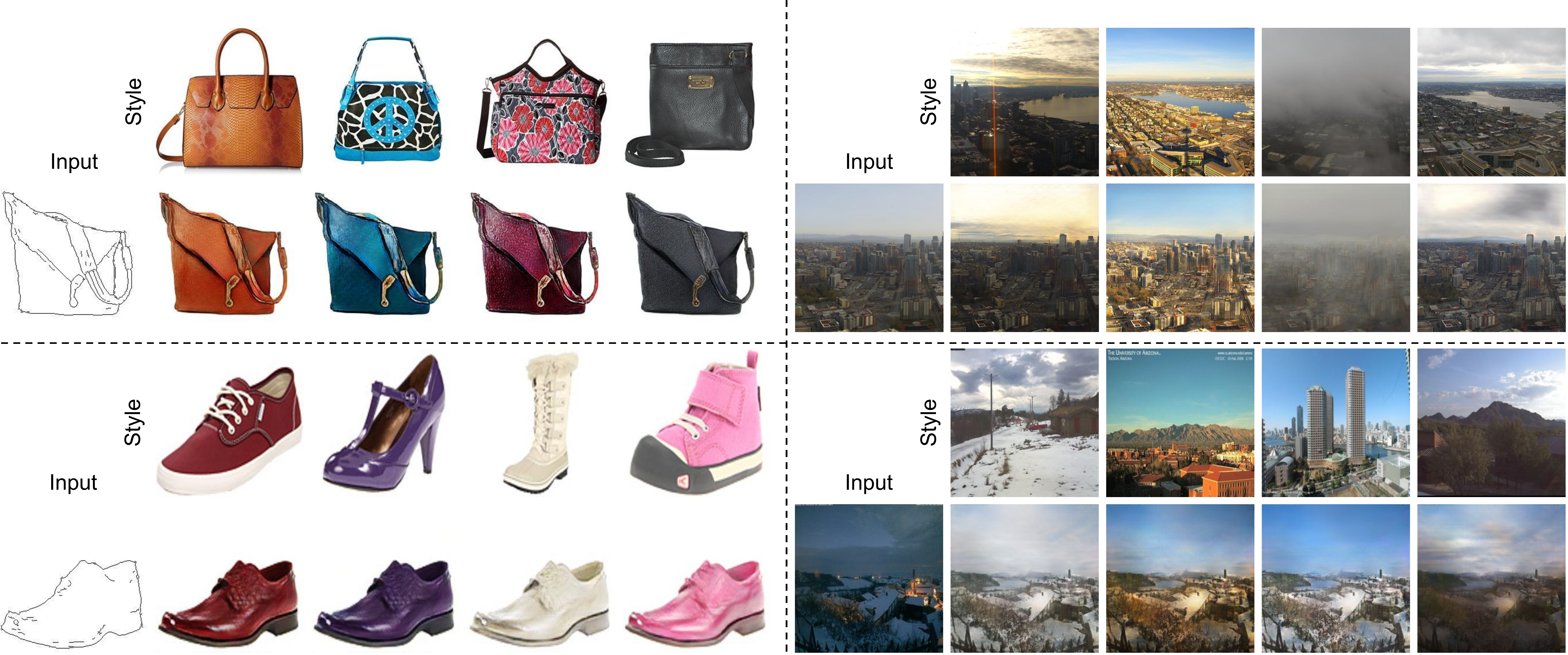}
    \vspace{-0.3cm}
    \caption{Style transfer for different datasets. We show output for applying different styles to each input image.}
    \vspace{-0.19in}
    \label{fig:app_transfer}
\end{figure*}

\subsection{Weakly-supervised encoder pre-training}
\label{sec:pretraining}

The goal of pre-training the encoder network $E$ is to learn a deterministic mapping from the style of a target image $I_i^B \in B$ to a latent style code $z_i = E(I_i^B)$.
Ideally, images with similar styles should be close in the style embedding space, while images with different styles should be far apart.
To supervise training such an embedding, we utilize the style loss~\cite{gatys2016image} as a distance metric to measure the style similarity between any two given images.
The style encoder network $E$ is then trained using a triplet loss~\cite{schroff2015facenet}, where the input is a triplet of images $(I_a, I_p, I_n)$, where $(I_a, I_p)$ have similar style, while $(I_a, I_n)$ have different style, as measured by the style loss metric.
The training objective for $E$ is given by:
\begin{equation} 
\begin{split}
\mathcal{L}^{\mbox{\scriptsize{tri}}}(I_a, I_{\mbox{\scriptsize{p}}}, I_{\mbox{\scriptsize{n}}}) = & \max\left(\left[\|z_a - z_p\|^2 - \|z_a - z_n\|^2 + \alpha \right], 0\right)\\
& + \lambda  \mathcal{L}^{\mbox{\scriptsize{reg}}}\left(z_a, z_p, z_n\right)
\end{split}
\end{equation}
where $\alpha$ is a separation margin, $\lambda$ is a relative weighting parameter between the main triplet objective and an optional regularization term $\mathcal{L}^{\mbox{\scriptsize{reg}}}(\cdot)$ which is an $L2$ regularization to encourage learning a compact latent space.
%
%
\begin{figure*}[t!]
    \centering
    \includegraphics[width=0.68\linewidth]{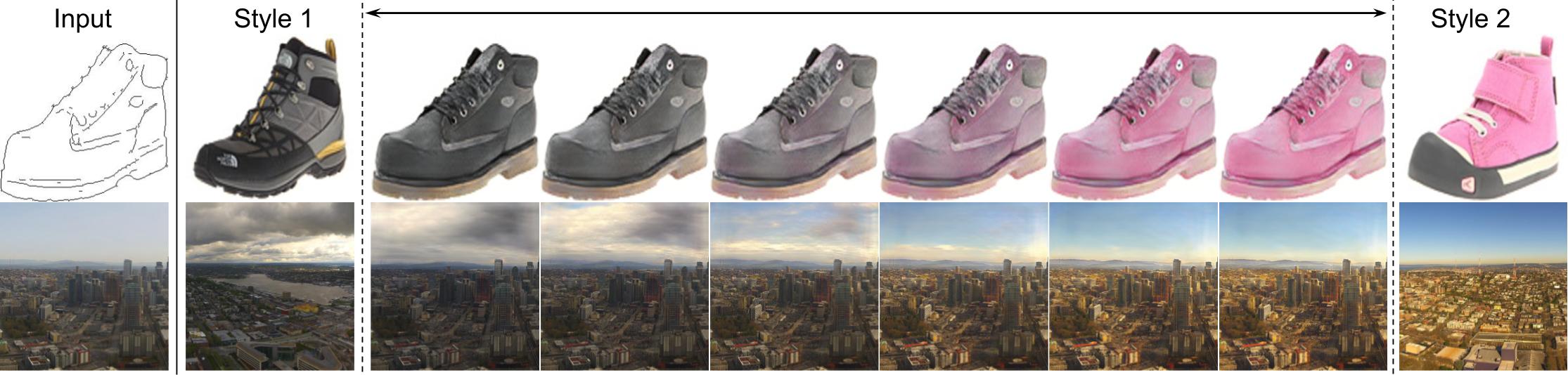}
    \vspace{-0.3cm}
    \caption{Style interpolation. Left column is the input to the generator $G$, second and last columns are input style images to the style encoder, and middle images are linear interpolation in the embedding space (figure better seen in zoom).}
    \vspace{-0.4cm}
    \label{fig:app_interpolation}
\end{figure*}

\noindent\textbf{Triplet selection.}
To generate triplets for pre-training the encoder $E$, we compute the set of $k_c$ closest and $k_f$ furthest neighbors for each anchor image $I_a$ as measured by the style loss. Then, for each anchor image $I_a$, we randomly sample a positive image $I_p$ and a negative image $I_n$ from the set of closest and furthest neighbors respectively.
We found that, for large datasets, it is sufficient to generate triplets for a subset of the training images.
%
One challenge is the set of images with an outlier style. Such images will be furthest neighbors to most images, and can mislead the training by just projecting outlier images to separate clusters.
To deal with this, we sample the negative style image $I_n$ from a larger set of furthest neighbors; while the positive image $I_p$ is sampled from a small set of closest neighbors so that it would have reasonable style similarity to the anchor image.

\subsection{Generator training}\label{sec:g_training}

After pre-training the style encoder $E$ (stage 1), we have established a mapping from images in the output domain, $I^B \in B$, to their style-embedding $z = E(I^B)$. Feeding the style embedding as input to the generator during training, the generator has good incentive to associate the style of output images to their corresponding style embedding instead of learning to hallucinate the style. It's important to retain the deterministic correspondence between images and their style codes to facilitate the job of the generator to discover this correlation. This is why, during stage 2, we keep the weights of the style encoder, $E$, fixed.
The forward pass reconstructs a training image $I^B_i$ as $\hat{I}^B_i = G(I^A_i, z_i)$, where $z_i = E(I^B_i)$.
The training objective is similar to that of pix2pix~\cite{isola2017image}:
\begin{equation}
    \mathcal{L}^{\mbox{\scriptsize{img}}}(I^B_i, \hat{I}^B_i) = \mathcal{L}_{\scriptsize{\text{cGAN}}}(I^B_i, \hat{I}^B_i) + \lambda_{\mbox{\scriptsize{rec}}} \mathcal{L}_{\mbox{\scriptsize{rec}}}(I^B_i, \hat{I}^B_i)
    \label{eqn:gen_loss}
\end{equation}
where we use the Least Square GAN loss (LSGAN)~\cite{mao2017least} for the $\mathcal{L}_{\scriptsize{\text{cGAN}}}$ term, and a VGG-based perceptual loss~\cite{johnson2016perceptual} for the reconstruction term $\mathcal{L}_{\mbox{\scriptsize{rec}}}$.
Once the generator has learned to associate the output style with the input style embedding, stage 3 finetunes both the generator, $G$, and the style encoder, $E$, together using the same objective~(\ref{eqn:gen_loss}).


%

\noindent\textbf{Style sampling.}
To perform multimodal synthesis on a given input at test time, we can capture the latent vector $z$ from any existing image and transfer the style to the generated image.
However, if we wish to sample styles directly from the latent distribution, one option is to enforce a prior on the latent distribution.
For example, we found it effective to add an $L2$ regularization on the latent vectors to enforce zero-mean embeddings and limit the variance of the latent space.
We then compute an empirical standard deviation for sampling.
Another alternative to enable sampling is to train a mapper network $\mathcal{M}$ to map the unit Gaussian to the latent distribution.
This can be done as a post-processing step after the style encoder has been trained and finetuned.
Specifically, we propose to train a mapper network $\mathcal{M}$ using the nearest-neighbor based Implicit Maximum Likelihood Estimation (IMLE) training~\cite{hoshen2019non,li2018implicit}.
The training objective is given by:
\begin{equation}
\resizebox{\linewidth}{!}{$
    \mathcal{M} = \argmin_{\tilde{\mathcal{M}}} \sum_i \|z_i - \tilde{\mathcal{M}}(e_i)\|_2^2, {\ }
    e_i = \argmin_{r_j} \|z_i - \tilde{\mathcal{M}}(r_j)\|_2^2
    \label{eqn:mapper_loss}
$}
\end{equation}
where $\{r_j\}$ is a set of random samples from the unit Gaussian prior, and for each latent code $z_i$, we select $e_i$ that generates the nearest neighbor $\mathcal{M}(e_i)$ to $z_i$.


\subsection{Generalizing the pre-training stage}
\label{sec:e_generalization}

The use of Gram matrices for Neural Style Transfer proved to be very effective and it reliably captures the style of arbitrary input images.
This implies that Gram matrices can reliably encode styles from a wide range of domains, and they are not specific to a certain domain.
Therefore, we hypothesize that encoder pre-training using a style-based triplet loss would learn a generic style embedding that can generalize across multiple domains and be effective for multi-modal I2I translation.
This would allow for performing the pre-training stage only once using auxiliary training data.
The finetuning stage eventually tweaks the embedding to better suit the specific target domain $B$.
We validate our hypothesis experimentally in $\S$\ref{sec:experiments}, and show that pre-training the style encoder on datasets other than the target domain $B$ doesn't degrade the performance. It can even improve the performance if the target dataset is small, in which case pre-training on an auxiliary dataset helps with the generalization of the overall model.
\section{Experimental evaluation}
\label{sec:experiments}

\begin{figure*}[t!]
    \centering
    \includegraphics[width=0.7\linewidth]{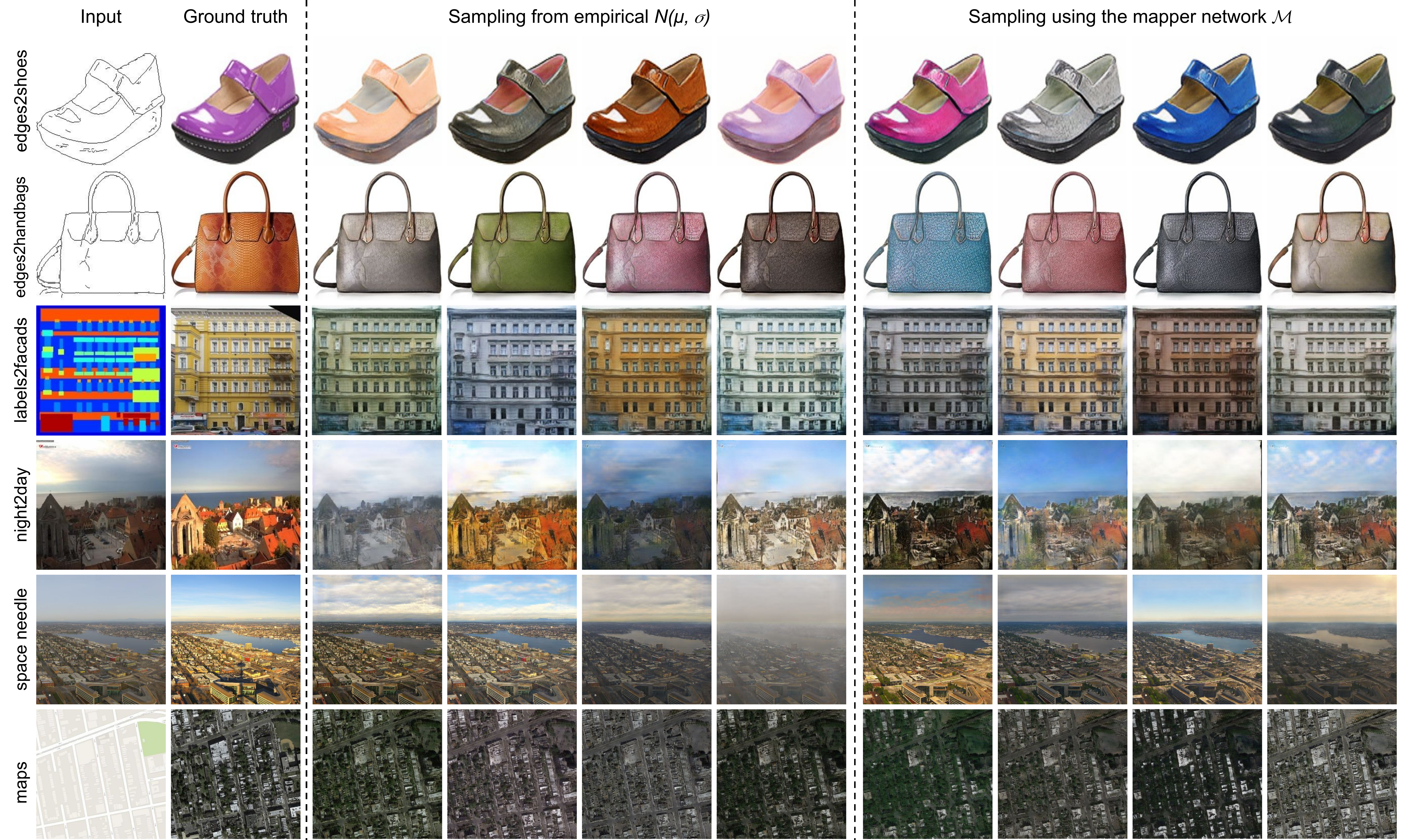}
    \vspace{-0.3cm}
    \caption{Style sampling for different datasets using our approach after full training (\emph{e.g.,} phase 3). We sample either from $N(\mu,\sigma)$, where $\mu, \sigma$ are computed from the train set (middle), or using the mapper network $\mathcal{M}$ (right).}
    \vspace{-0.1in}
    \label{fig:app_sampling}
\end{figure*}


\noindent
\textbf{Datasets.}
We evaluate our approach on five standard I2I translation benchmarks used in~\cite{isola2017image,zhu2017toward}; Architectural labels $\rightarrow$ photo, aerial $\rightarrow$ map, edges $\rightarrow$ shoes/handbags and night $\rightarrow$ day.
In addition, we use the Space Needle timelapse dataset~\cite{spaceneedle}, which consists of 2068 paired images with a $8280 \times 1080$ resolution, where the input domain includes images with temporally smoothed appearance, and the output domain contains real images spanning different lighting and weather conditions.

\noindent
\textbf{Baselines.}
\change{
While we report numbers for retrained models using the official code released with BicycleGAN (BicycleGAN v0) for completeness, we mainly compare to two stronger baselines:
\begin{itemize}[leftmargin=*]
    \item 
        \textbf{BicycleGAN v1}: we implement BicycleGAN using the same network architecture as used in our approach to have a fair comparison (see supp.~material for implementation details).
    
    \item 
        \textbf{MUNIT-p}:
        We train MUNIT~\cite{huang2018multimodal} in a \emph{paired} setup by applying its cross-cycle consistency constraint as follows: the input is a pair of training examples $(I_1^A, I_1^B), (I_2^A, I_2^B)$ for which we obtain their respective style embeddings $z_1 = E(I_1^B), z_2 = E(I_2^B)$.
    We then apply a 2-step cyclic reconstruction of $I_1^B, I_2^B$; in the first step, we generate both images with a swapped style $u = G(I^A_1, z_2), v = G(I^A_2, z_1)$.
    In the second step, we re-capture the latent style vectors $\hat{z}_2 = E(u), \hat{z}_1 = E(v)$ and generate the original images $I_1^B, I_2^B$ by swapping the style again: $\hat{I}_1^B = G(I^A_1, \hat{z}_1)$, $\hat{I}_2^B = G(I^A_2, \hat{z}_2)$.
    We add a cyclic reconstruction loss for $\hat{I}_1^B, \hat{I}_2^B$. 
\end{itemize}
}

\begin{table*}[t!]
  \caption{Validation set reconstruction quality, as measured by \emph{PSNR} (higher is better) and \emph{LPIPS}~\cite{zhang2018unreasonable} (lower is better), for various datasets.
  We compare between retraining BicycleGAN~\cite{zhu2017toward} authors' released code ({Bicycle v0}), the baselines ({BicycleGAN v1 and MUNIT-p}) described in $\S$\ref{sec:experiments}, and our approach both before finetuning ({ours - stage 2}), and after finetuning ({ours - stage 3}).}
  \label{table:rec}
  \vspace{-0.25cm}
  \centering
  \renewcommand{\arraystretch}{1.2}
  \renewcommand{\tabcolsep}{1.5mm}
  \resizebox{0.88\linewidth}{!}{
  \begin{footnotesize}
      \begin{tabular}{@{}lllllllllllll@{}}
        \toprule
        \multirow{2}{*}{}  & \multicolumn{2}{c}{edges2handbags} & \multicolumn{2}{c}{edges2shoes} & \multicolumn{2}{c}{labels2facades} & \multicolumn{2}{c}{night2day} & \multicolumn{2}{c}{maps} & \multicolumn{2}{c}{space needle} \\
         & PSNR $\uparrow$ & LPIPS $\downarrow$ & PSNR $\uparrow$ & LPIPS $\downarrow$ & PSNR $\uparrow$ & LPIPS $\downarrow$ & PSNR $\uparrow$ & LPIPS $\downarrow$ & PSNR $\uparrow$ & LPIPS $\downarrow$ & PSNR $\uparrow$ & LPIPS $\downarrow$ \\ 
         \midrule 
        {Bicycle v0}  & 17.08 & 0.255 & 20.24 & 0.177 & 12.64 & 0.431 & 13.25 & 0.520 & 14.32 & 0.396 & -- & -- \\
        \midrule
        {Bicycle v1} & 18.52 & 0.198 & 21.84 & 0.124 & 13.08 & 0.378 & 13.88 & 0.491 & 14.67 & 0.359 & 19.72 & 0.233 \\
        {MUNIT-p} & \textbf{19.23} & 0.192 & 22.51 & 0.132 & 13.36 & 0.375 & 14.48 & 0.480 & \textbf{16.17} & 0.407 & 19.84 & 0.238 \\ 
        \midrule
        {\textbf{Ours} - stage 2} & 18.01 & 0.209 & 21.40 & 0.140 & \textbf{13.44} & 0.383 & 14.34 & \textbf{0.476} & 15.08 & 0.392 & \textbf{21.39} & \textbf{0.227} \\
        {\textbf{Ours} - stage 3} & 18.91 & \textbf{0.177} & \textbf{22.68} & \textbf{0.117} & \textbf{13.44} & \textbf{0.370} & \textbf{15.05} & \textbf{0.452} & 15.15 & \textbf{0.349} & \textbf{22.11} & \textbf{0.187}\\
        \bottomrule
      \end{tabular}
  \end{footnotesize}
  }
\vspace{-0.17in}
\end{table*}

\subsection{Image reconstruction}
We report the reconstruction quality of validation set images, using both \emph{PSNR} and \emph{AlexNet}-based \emph{LPIPS}~\cite{zhang2018unreasonable} metrics, in Table~\ref{table:rec}.
Note that our results with the pre-trained embeddings without finetuning (stage 2) are on-par-with the baselines.
This verifies the validity of our approach and that style-based encoder pre-training successfully learns to distinguish different modes in the output domain, which proves effective for training multi-modal I2I networks.
Finetuning (stage 3) further improves our results compared to the baselines.
See supp.~material for more quantitative comparison including Inception Scores (\emph{IS}).
Figure~\ref{fig:qualitative} shows qualitatively how our approach reconstructs the target style more faithfully.
Our approach not only matches the ground truth colors better, but also texture (e.g., left column, first and third rows).
While the baselines rely on VAEs to provide the style sampling property, we observe that, for a low dimensional latent space, the noise robustness added by VAEs reduces the expressiveness of the latent space, since slight changes to one style code would still be mapped to the same point in the latent space.
This explains why our approach, which doesn't use VAEs, achieves more faithful style capture and reconstruction.
We verified this by studying the effect of removing the VAE component from the baselines, which improved their performance as shown in $\S$\ref{sec:ablation}.
%


\subsection{Style transfer and sampling}
Figure~\ref{fig:app_transfer} shows style transfer to images from the validation set of different datasets.
Note how the style transfer copies the weather conditions in the Space Needle and Night2day datasets.
For example, in the Space Needle dataset, we show sunset, sunny, foggy and cloudy weather.
Also, the Night2day examples exhibit variation in lighting conditions including transferring whether the surface is sunlit or not, as well as different cloud patterns and clear skies.
Comparison to the baselines' style transfer results in the supplementary material further highlights the improvements of our approach.
We can also sample random styles directly from the latent distribution as described in $\S$\ref{sec:g_training}.
Figure~\ref{fig:app_sampling} shows results for both adhoc sampling from the assumed $N(\mu, \sigma)$ empirical distribution, as well as formally sampling from a unit Gaussian using the mapper network $\mathcal{M}$.
Note that the diversity of sampled styles doesn't stem from simple color changes;
for example, sampled styles for the Space Needle dataset show clear weather changes, such as cloudy/sunny, different cloud patterns and even sampling foggy weather which was present in some images in the training set.
In the Maps dataset, the existence and/or density of bushes clearly varies between different sampled styles.
Also, in the Edges2handbags dataset, the texture of the bag varies between smooth and rough leather (better seen in zoom).
While the results of both sampling methods look good, we note that the assumption for adhoc sampling is not explicitly enforced, and thus could lead to sampling bad style codes outside the distribution (see supp.~material for examples). 
\subsection{Style interpolation}
Figure~\ref{fig:app_interpolation} shows style interpolation by linearly interpolating between two latent vectors.
For example, note the smooth change in lighting and cloud patterns when going from cloudy to sunny in the Space Needle dataset.
More interpolation results on other datasets, and comparison with interpolation results of the baselines can be found in the supplementary material.
%

\begin{table*}[ht!]
\centering
\begin{footnotesize}
\caption{Ablation study of the effect of different components and loss terms using the edges2handbags dataset. We study direct and cyclic reconstructions on ground truth images (dir\_recon, cyc\_recon), discriminator loss on direct reconstructions and on generated images with a randomly sampled style (D\_dir, D\_rand\_z), latent reconstruction (z\_recon), $L2$ and {KL} regularization on the latent vector $z$ (z\_L2, z\_KL), and finally the use of VAE vs.\ just an auto-encoder.}
\label{table:ablation_losses}
\vspace{-0.2cm}
\centering
\renewcommand{\arraystretch}{1.2}
\renewcommand{\tabcolsep}{1.5mm}
\resizebox{0.9\linewidth}{!}{
    \begin{tabular}{@{}lccccccccccc@{}}
    \toprule
    \multirow{2}{*}{Approach} & \multicolumn{8}{c}{Loss setup} & \multirow{2}{*}{IS$\uparrow$} & \multirow{2}{*}{PSNR$\uparrow$} & \multirow{2}{*}{LPIPS $\downarrow$} \\ \cmidrule{2-9}
                   & dir\_recon & cyc\_recon & D\_dir & D\_rand\_z & z\_recon & z\_L2  & z\_KL   & VAE    &                  &                 &  \\ \midrule
    {Bicycle v1}   & \cmark     & --         & \cmark & \cmark     & \cmark   & --     & \cmark  & \cmark & $2.31 \pm 0.05$ & $18.28 \pm 0.30$ & $0.201 \pm 0.003$ \\ 
    {MUNIT-p}   & \cmark     & \cmark     & \cmark & \cmark     & \cmark   & --     & \cmark  & \cmark & $2.45 \pm 0.07$ & $18.96 \pm 0.30$ & $0.192 \pm 0.002$ \\ \midrule 
    {Bicycle v2} & \cmark     & --         & \cmark & \cmark     & \cmark   & --     & \cmark  & --     & $2.36 \pm 0.12$ & $19.02 \pm 0.10$ & $0.175 \pm 0.001$ \\ 
    {MUNIT-p v2} & \cmark     & \cmark     & \cmark & \cmark     & \cmark   & --     & \cmark  & --     & $2.44 \pm 0.06$ & $19.34 \pm 0.07$ & $0.176 \pm 0.002$ \\ \midrule 
    {Bicycle v3} & \cmark     & --         & \cmark & \cmark     & \cmark   & \cmark & --      & --     & $2.34 \pm 0.08$ & $19.21 \pm 0.06$ & $0.177 \pm 0.002$ \\ 
    {MUNIT-p v3} & \cmark     & \cmark     & \cmark & \cmark     & \cmark   & \cmark & --      & --     & $2.33 \pm 0.04$ & $19.24 \pm 0.09$ & $0.180 \pm 0.001$ \\ \midrule 
    {\textbf{Ours} v1}      & \cmark     & \cmark     & \cmark & \cmark     & \cmark   & \cmark & --      & --     & $2.41 \pm 0.07$ & $18.97 \pm 0.13$ & $0.189 \pm 0.004$ \\ 
    {\textbf{Ours} v2}      & \cmark     & --         & \cmark & \cmark     & \cmark   & \cmark & --      & --     & $2.43 \pm 0.03$ & $18.94 \pm 0.10$ & $0.183 \pm 0.002$ \\ 
    {\textbf{Ours} v3}      & \cmark     & --         & \cmark & --         & --       & \cmark & --      & --     & $2.42 \pm 0.03$ & $18.94 \pm 0.05$ & $0.176 \pm 0.001$ \\ 
    {\textbf{Ours} v4}      & \cmark     & --         & \cmark & --         & --       & --     & --      & --     & $2.46 \pm 0.03$ & $18.94 \pm 0.02$ & $0.177 \pm 0.001$ \\ \bottomrule 
    \end{tabular}
}
\end{footnotesize}
\vspace{-0.3cm}
\end{table*}

\begin{table*}[!ht]
\centering
    \begin{minipage}{.5\linewidth}
        \caption{Generalization of a pretrained style encoder $E$. We report validation set reconstruction for the edges2handbags and night2day datasets when pretraining with different datasets. Stages 2, 3 show results before/after finetuning $E$ respectively.}
        \label{table:generalize_enc}
        \centering
        \vspace{-0.1in}
        \resizebox{\linewidth}{!}{
            \begin{tabular}{@{}llcccc@{}}
            \toprule 
            \multirow{2}{*}{Dataset} & \multirow{2}{*}{pretrain dataset} & \multicolumn{2}{c}{Stage 2} & \multicolumn{2}{c}{Stage 3} \\ \cmidrule{3-6}
            & & PSNR $\uparrow$ & LPIPS $\downarrow$ & PSNR $\uparrow$ & LPIPS $\downarrow$ \\ \midrule
            \multirow{3}{*}{edges2handbags} & edges2handbags & {18.01} & {0.209} & 18.91 & 0.177 \\ 
            & edges2shoes & 17.89 & 0.215 & 18.96 & 0.176 \\ 
            & space\_needle  & 17.86 & 0.221 & {19.02} & {0.175} \\ \midrule 
            \multirow{3}{*}{night2day} & night2day & 13.75 & 0.489 & {15.15} & 0.454 \\ 
            & space\_needle & {14.34} & {0.476} & 15.05 & {0.452} \\ 
            & edges2handbags & 13.91 & 0.492 & 15.03 & 0.461 \\ \bottomrule
            \end{tabular}
        }
    \end{minipage}%
    ~ \quad
    \begin{minipage}{.45\linewidth}
    \caption{Diversity score is the average LPIPS distance~\cite{zhang2018unreasonable}. User preference score is the percentage a method is preferred over `Ours v4', on the edges2shoes dataset.}
      \label{table:diversity}
       \vspace{-0.1in}
      \centering
      \resizebox{0.78\linewidth}{!}{
        \begin{tabular}{@{}lccc@{}}
        \toprule 
        \multirow{2}{*}{Approach} & \multirow{2}{*}{\parbox{1.5cm}{\centering LPIPS $\uparrow$\\(transfer)}} & \multirow{2}{*}{\parbox{1.5cm}{\centering LPIPS $\uparrow$ \\(sampling)}} & \multirow{2}{*}{\parbox{1.9cm}{\centering User \\preference $\uparrow$}}\\ 
        &&& \\ \midrule
        {Bicycle v1} & $0.102$ & $0.119$ & $30.0\%$ \\
        {MUNIT-p} & $0.138$ & $0.132$ & $37.7\%$ \\ \midrule 
        {\textbf{Ours} v1} & $0.153$ & $0.148$ & $46.5\%$ \\ 
        {\textbf{Ours} v2} & {0.171} & $0.140$ & $41.1\%$ \\ 
        {\textbf{Ours} v3} & $0.149$ & {0.165} & {50.4\%} \\ 
        {\textbf{Ours} v4} & $0.154$ & $0.132$ & $50\%$ \\ \bottomrule 
        \end{tabular}
    }
    \end{minipage} 
\vspace{-0.1in}
\end{table*}

\subsection{Pre-training generalization}
Since the notion of style, as defined in the Neural Style Transfer literature, is universal and not specific to a certain domain, we hypothesized that style-based encoder pre-training would learn a generic style embedding that can generalize across multiple domains and be effective for multi-modal image I2I translation.
Here, we experimentally verify our hypothesis in Table~\ref{table:generalize_enc}.
For a target dataset, we train the generator $G$ three times, each with different pre-training of the style encoder $E$: 
(1) same dataset pre-training: pre-train $E$ using the output domain $B$ of the target dataset.
(2) similar-domain pre-training: pre-train on a different dataset, but whose output domain bears resemblance to the output domain of the target dataset (e.g., edges2shoes and edges2handbags, or day images from night2day and the Space Needle timelapse dataset).
(3) different-domain pre-training: pre-train on a different dataset whose output domain has different styles from that of the target dataset (e.g., edges2handbags and the Space Needle timelapse datasets, or night2day and edges2handbags datasets).
Table~\ref{table:generalize_enc} shows that without finetuning (i.e., stage 2), the edges2handbags dataset shows a slight performance degradation when going from pre-training on the same dataset, to pre-training on a similar-domain dataset, and finally pre-training on a different-domain dataset. 
On the other hand, the night2day dataset has only $\sim$100 unique scenes for training. So, pre-training on another dataset such as Space Needle generalizes better to new scenes in the validation set, since it helps avoid overfitting the small number of unique scenes in the training set.
After finetuning, performance differences further reduce to be insignificant.
%
We also investigate the generalization of the proposed encoder pre-training to the case of using non-style distance metrics in the supplementary material.
%

\begin{figure*}[ht!]
    \centering
    \vspace{-.2cm}
    \includegraphics[width=0.62\linewidth]{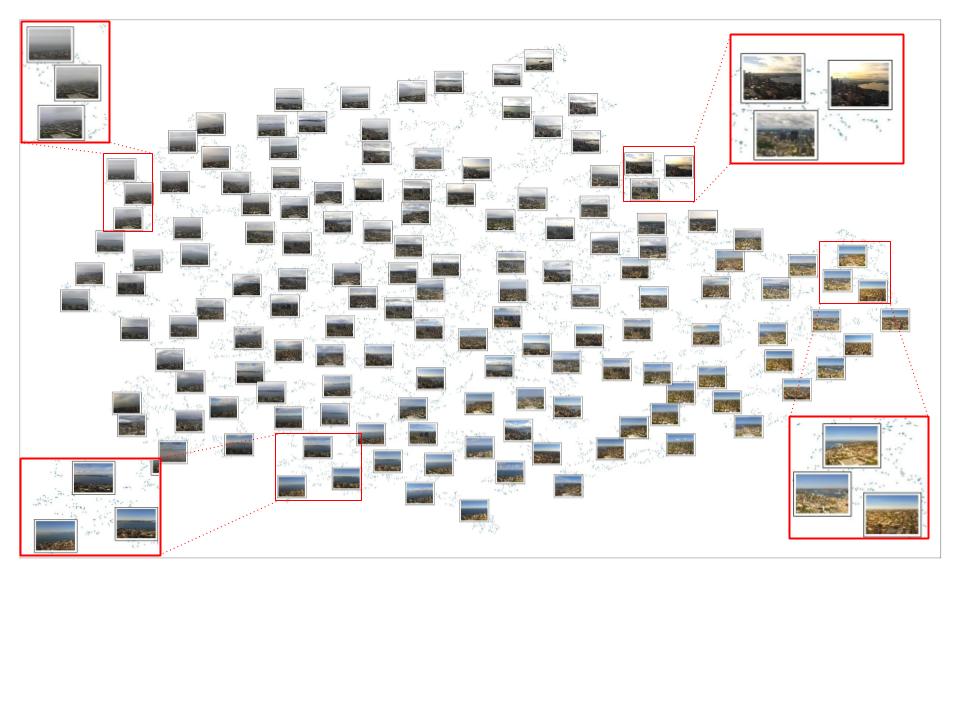}
    \vspace{-2cm}
    \caption{t-SNE plot for the pre-trained latent space of the Space Needle timelapse dataset. Images with similar styles (\emph{e.g.,}~ weather conditions and lighting) lie closely in the latent space. (figure best seen in zoom)}
    \label{fig:pretrained_latent_vis}
\vspace{-0.3cm}
\end{figure*}

%
\subsection{Ablative study}
\label{sec:ablation}
We investigate the role of different loss terms as we transition from the loss setup of the baselines to that of our training approach.
We first remove the variational part in both BicycleGAN v1 and MUNIT-p baselines resulting in Bicycle v2, MUNIT-p v2.
We further remove the Gaussian prior and replace the KL loss with an L2 regularization in Bicycle v3, MUNIT-p v3.
To maintain random latent vector sampling during training without a prior, we sample a random training image, and use its style code.
We define different versions of our approach (v1, v2, v3, and v4) based on different loss setup during training as follows:
we start with `Ours v1', which has the same setup as MUNIT-p v3, except that it uses pre-trained embeddings as described in $\S$\ref{sec:pretraining}.
We then remove cyclic reconstruction, random z sampling, and L2 regularization terms resulting in `Ours v2', `v3', and `v4' respectively.
We run each setup on the edges2handbags dataset.
In order to draw more reliable conclusions, we repeat each experiment 3 times and report the mean and standard deviation in Table~\ref{table:ablation_losses}.
%
We notice that removing the variational part in VAEs is enough to improve the reconstruction results.
While VAEs in general are robust to noise in the input latent, we observe that this comes at the expense of the expressiveness of the latent space (e.g., less faithful style capture and transfer), especially for low dimensional latents.
%
We also observe that our approach generally performs better with less constraints (loss terms). 
For example, ``Ours v1, v2'' have lower results than their ``Bicycle v3'', ``MUNIT-p v3'' counterparts.
This shows that the main benefit of pre-trained embeddings is when the network is less constrained.



\subsection{Diversity and user study}
We evaluate diversity by computing the average LPIPS distance over 1600 outputs.
We measure diversity on two setups: we sample 100 validation images, and (1) apply style transfer from 16 randomly sampled images, or (2) we sample 16 random codes using the mapper network $\mathcal{M}$ to obtain 1600 outputs.
%
We also measure the realism and faithfulness of style transfer through a user study, where 30 participants are shown an input shoe sketch, an input style image and two style transfer outputs.
They are asked to choose which output looks more realistic, and if both are realistic, then which transfers the style more faithfully.
We fix `Ours v4' approach as anchor and compare other methods to it. Table~\ref{table:diversity} shows that the baselines achieve lower diversity and user preference compared to our approach, specially in the style transfer setup.
Different variations of our method, except for `Ours v2' yield similar diversity and user preference scores.
%
\change{
We observe that `Ours v2' shows artifacts in some outputs, leading to higher diversity but lower user preference.
Our diversity results for the style sampling setup have some variation and are sensitive to the mapper network training, but are still either on-par or better than the baselines.
}

\subsection{Visualizing pre-trained embeddings}
\label{sec:pretrained_latent_vis}
Figure~\ref{fig:pretrained_latent_vis} visualizes the pre-trained latent space learned by the style encoder $E$.
The visualization shows meaningful clusters of similar styles (weather conditions for the Space Needle timelapse dataset).
Refer to supp.~ material for visualization of the latent space after finetuning, as well as the latent space learned by the baselines.


\noindent
\textbf{More analysis and discussion:}
Please refer to the supplementary material for convergence analysis, training time comparison, and more quantitative and qualitative results.
\section{Conclusion}
We investigated the effectiveness of Style-based Encoder Pre-training (StEP) for the task of multi-modal I2I translation. 
The proposed pre-training can be done once on auxiliary data and generalizes well to several domains.
This allows for a faster training of I2I translation networks with fewer losses and achieves more faithful style capture and transfer.
Furthermore, we studied the contribution of different loss terms to the task of multi-modal I2I translation, where we discovered that noise added by a variational auto-encoder can limit the expressiveness of low-dimensional latent spaces.
We proposed two simple alternatives to VAEs to provide latent code sampling.
Finally, we achieved state-of-the-art results on several benchmarks.


\bigskip
{\footnotesize \noindent\textbf{Acknowledgements.} We would like to thank Ricardo Martin-Brualla for helping with initial concepts and reviewing drafts. We also thank Kamal Gupta and Alex Hanson for suggesting the acronym for our approach, and the members of the Perception and Intelligence (PI) Lab for their helpful feedback. The project was partially funded by DARPA MediFor (FA87501620191) and DARPA SAIL-ON (W911NF2020009) programs.}

\clearpage


%
%

{\small
\bibliographystyle{ieee_fullname}
\bibliography{references.bib}
}

\clearpage
\appendix
\section{Appendix}
\label{sec:appendix}
\subsection{Implementation details}
Triplet selection requires computing the set of nearest and furthest neighbors to each anchor image. 
When pre-training using a large dataset, we found it sufficient to randomly sample a subset of 8000 images and sample triplets from this subset.
This number was chosen to ensure fast nearest-neighbor computation.

The generator network $G$ has a symmetric encoder-decoder architecture based on~\cite{wang2017high}, with extra skip connections by concatenating feature maps of the encoder and decoder.
We use a multiscale-patchGAN discriminator~\cite{wang2017high} with 3 scales and employ a LSGAN~\cite{mao2017least} loss.
The mapper network $\mathcal{M}$ is a multi-layer perceptron (MLP) with three 128-dimensional hidden layers and a \emph{tanh} activation function.
For the reconstruction loss, we use the perceptual loss~\cite{johnson2016perceptual} evaluated at $\texttt{conv}_{i,2}$ for $i \in [1, 5]$ of VGG~\cite{simonyan2014very} with linear weights of $w_i = 1/2^{6-i}$ for $i \in [1, 5]$.
The architecture of the style encoder $E$ is adopted from~\cite{lee2018diverse}, and we use a latent style vector $z \in \mathbb{R}^8$.
Our optimizers setup is similar to that in~\cite{zhu2017toward}. We use three \emph{Adam} optimizers: one for the generator $G$ and encoder $E$, another for the discriminator $D$, and another optimizer for the generator $G$ alone with $\beta_1=0, \beta_2=0.99$ for the three optimizers, and learning rates of $0.001, 0.001$ and $0.0001$ respectively. We use a separate \emph{Adam} optimizer for the mapper network $\mathcal{M}$ with $\beta_1=0.5, \beta_2=0.99$, and a learning rate of $0.01$ with a decay rate of 0.7 applied every 50 steps.
Relative weights for the loss terms are $\lambda_{\mbox{\scriptsize{cGAN}}}=1$, $\lambda_{\mbox{\scriptsize{rec}}}=0.02$ and $\lambda_{\mbox{\scriptsize{$L2$}}}=0.01$ for the GAN loss, reconstruction loss, and $L2$ latent vector regularization respectively. When sampling triplets for any anchor image $I_c$, we use $k_c=5, k_f=13$ for the size of the set of close and far neighbors respectively.
For more implementation details, refer to our code \url{http://www.cs.umd.edu/~mmeshry/projects/step/}.

\subsection{Training time} 
Simplifying the training objective allows for faster training, as well as a larger batch size due to lower memory usage.
Table~\ref{table:runtime} shows the processing time per 1000 training images for the baselines as well as different variations of our approach as defined in Table 2 in the main text.

\begin{table}[!ht]
\centering
      \caption{Training time (in seconds) per 1000 images for the baselines, as well as different versions of our approach (defined in Table 2 in the main text).} 
      \label{table:runtime}
        \centering
        \resizebox{\linewidth}{!}{
            \begin{tabular}{@{}lcccc@{}}
            \toprule 
            \multirow{2}{*}{Approach} & \multirow{2}{*}{\parbox{0.9cm}{\centering Batch\\size}} & \multirow{2}{*}{\parbox{1.5cm}{\centering time/kimg$\downarrow$\\(sec)}} & \multirow{2}{*}{\parbox{1.4cm}{\centering Max batch size}} & \multirow{2}{*}{\parbox{1.5cm}{\centering time/kimg$\downarrow$\\(sec)}} \\ 
            &&&& \\ \midrule
            {Bicycle v1} & 8 & 93.11 & 12 & 85.36 \\
            {MUNIT-p} & 8 & 155.72 & 8 & 155.72 \\ \midrule 
            {\textbf{Ours} v1} & 8 & 145.50 & 8 & 145.50 \\ 
            {\textbf{Ours} v2} & 8 & 98.55 & 12 & 93.04 \\ 
            {\textbf{Ours} v3} & 8 & 64.92 & 16 & 53.80 \\ \bottomrule 
            \\
            \end{tabular}
        }

\vspace{-0.3in}
\end{table}


\begin{figure}[ht!]
    \centering
    \includegraphics[width=0.85\linewidth]{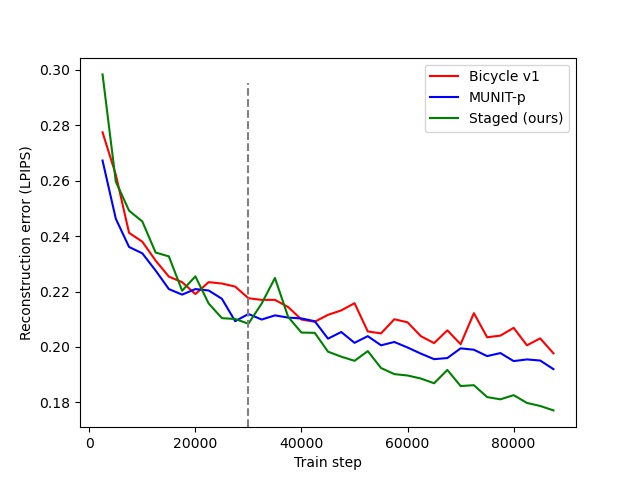}
    \caption{Convergence comparison between the proposed staged training (ours - v3) and the BicycleGAN baselines measured by the reconstruction error (LPIPS) of the validation set of the edges2handbags dataset. Dotted line shows the transition between stages 2 and 3 of our training (i.e, switching from a fixed $E$ to finetuning both $G$ and $E$ together).}
    \label{fig:convergence}
\end{figure}

\begin{figure*}[t!]
    \centering
    \includegraphics[width=0.86\linewidth]{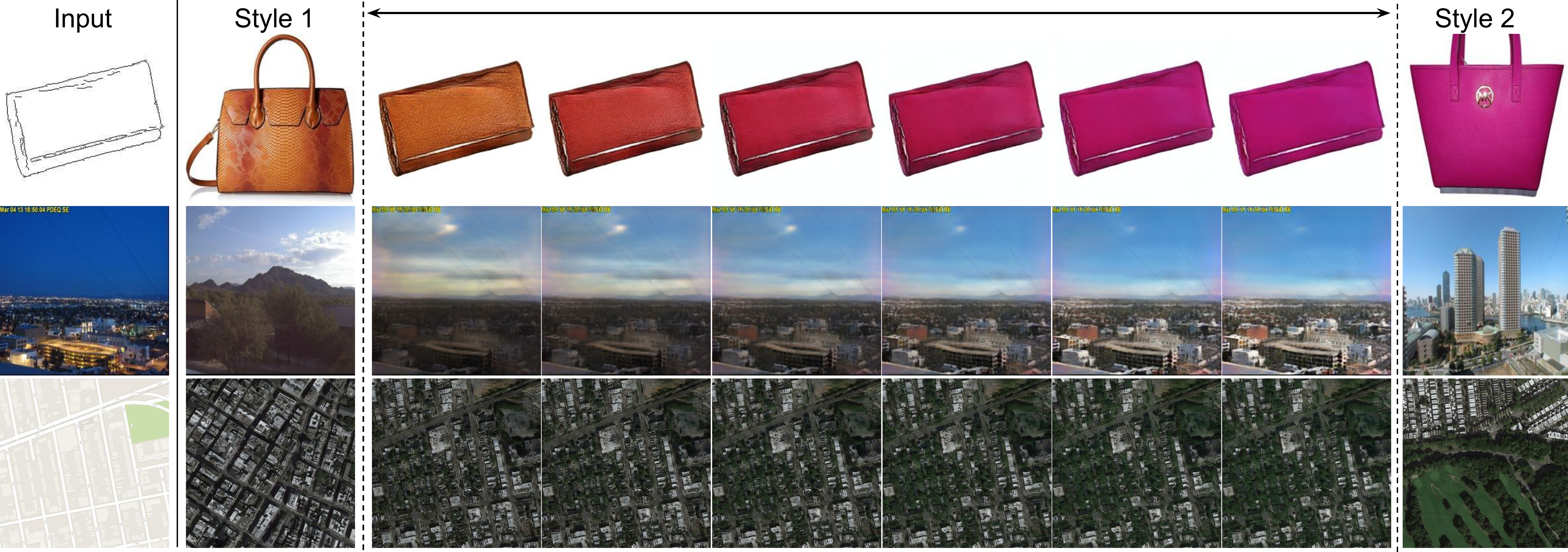}
    \caption{Style interpolation. Left column is the input to the generator $G$, second and last columns are input style images to the style encoder, and middle images are linear interpolation in the embedding space (figure better seen in zoom).}
    \label{fig:app_interpolation_more}
\end{figure*}

\begin{figure*}[t!]
    \centering
    \includegraphics[width=0.86\linewidth]{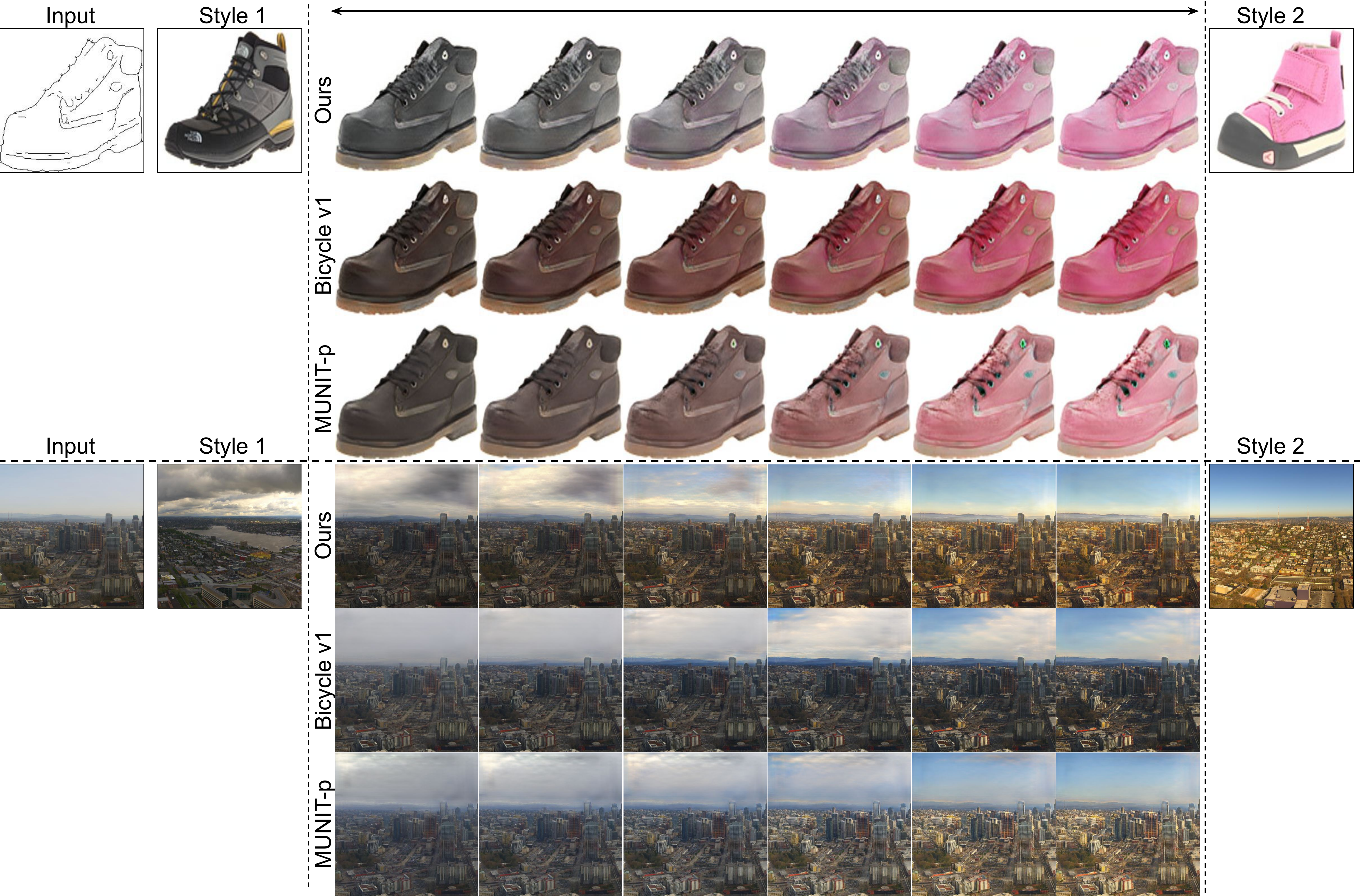}
    \vspace{-0.3cm}
    \caption{Style interpolation. Left column is the input to the generator $G$, second and last columns are input style images to the style encoder, and middle images are linear interpolation in the embedding space (figure better seen in zoom).}
    \vspace{-0.4cm}
    \label{fig:app_interpolation_cmp}
\end{figure*}

\begin{figure*}[t!]
    \centering
    \includegraphics[width=0.76\linewidth]{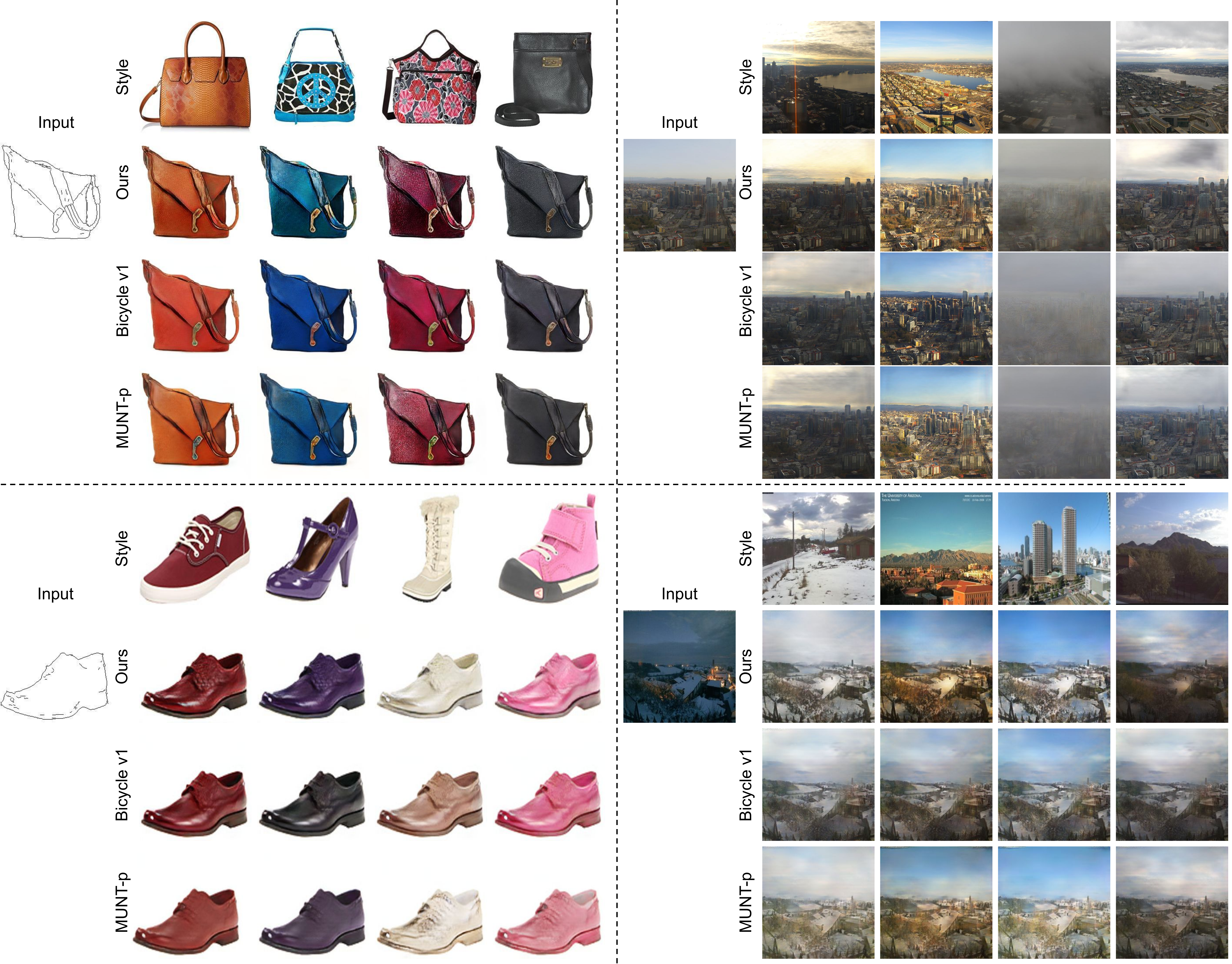}
    \caption{Style transfer comparison on different datasets. For each dataset, we apply different styles to the same input image and show the output of different methods.}
    \label{fig:style_transfer_cmp}
\end{figure*}

\begin{figure*}[ht!]
    \centering
    \captionsetup[subfigure]{aboveskip=1pt,belowskip=3pt}
    \begin{subfigure}[b]{0.63\linewidth}
        \centering
        \includegraphics[width=\linewidth]{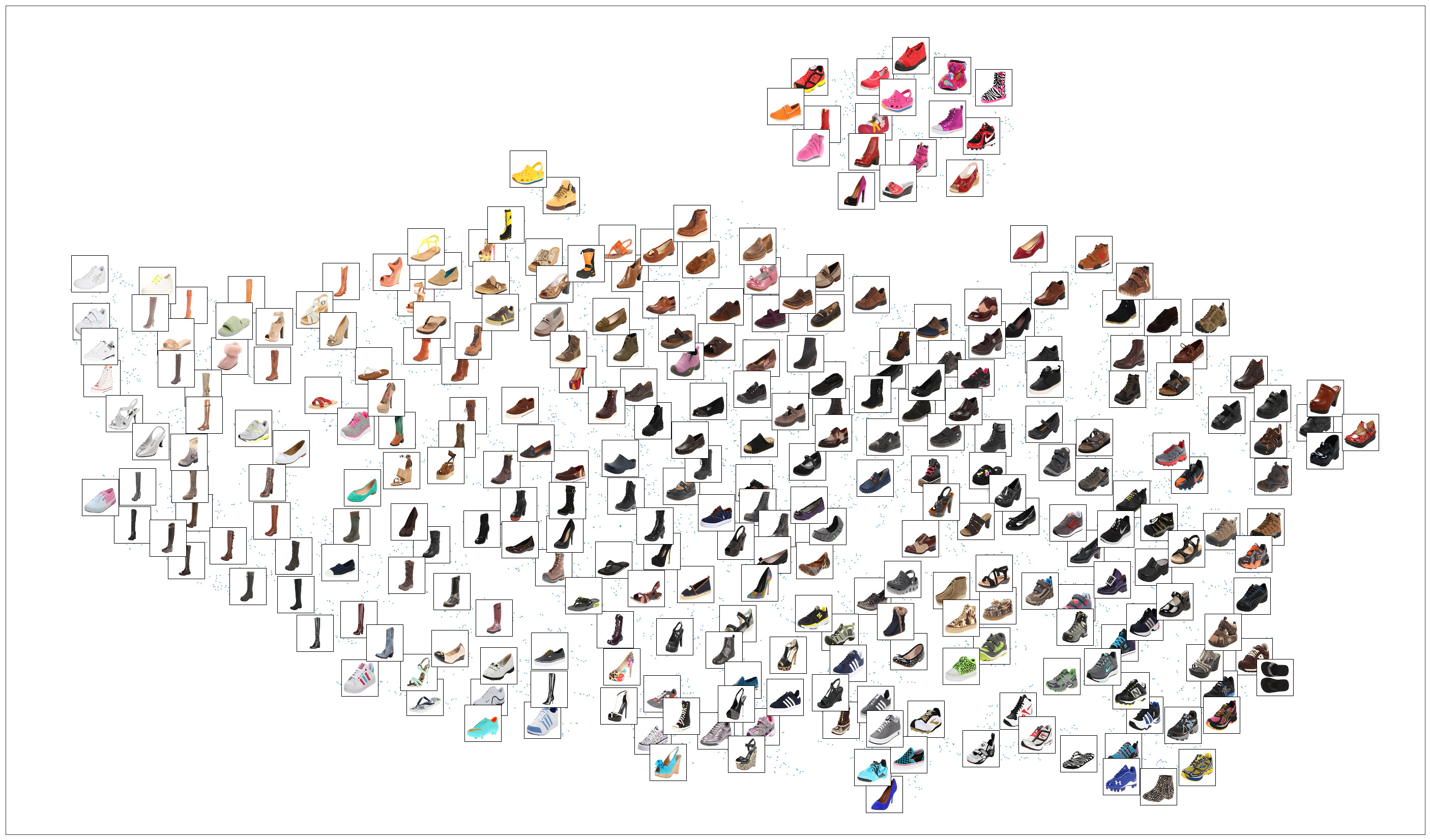}\\
        \caption{Our approach: after style pretraining.}
        \label{fig:latent_vis_pretrain}
    \end{subfigure}
    \begin{subfigure}[b]{0.63\linewidth}
        \centering
        \includegraphics[width=\linewidth]{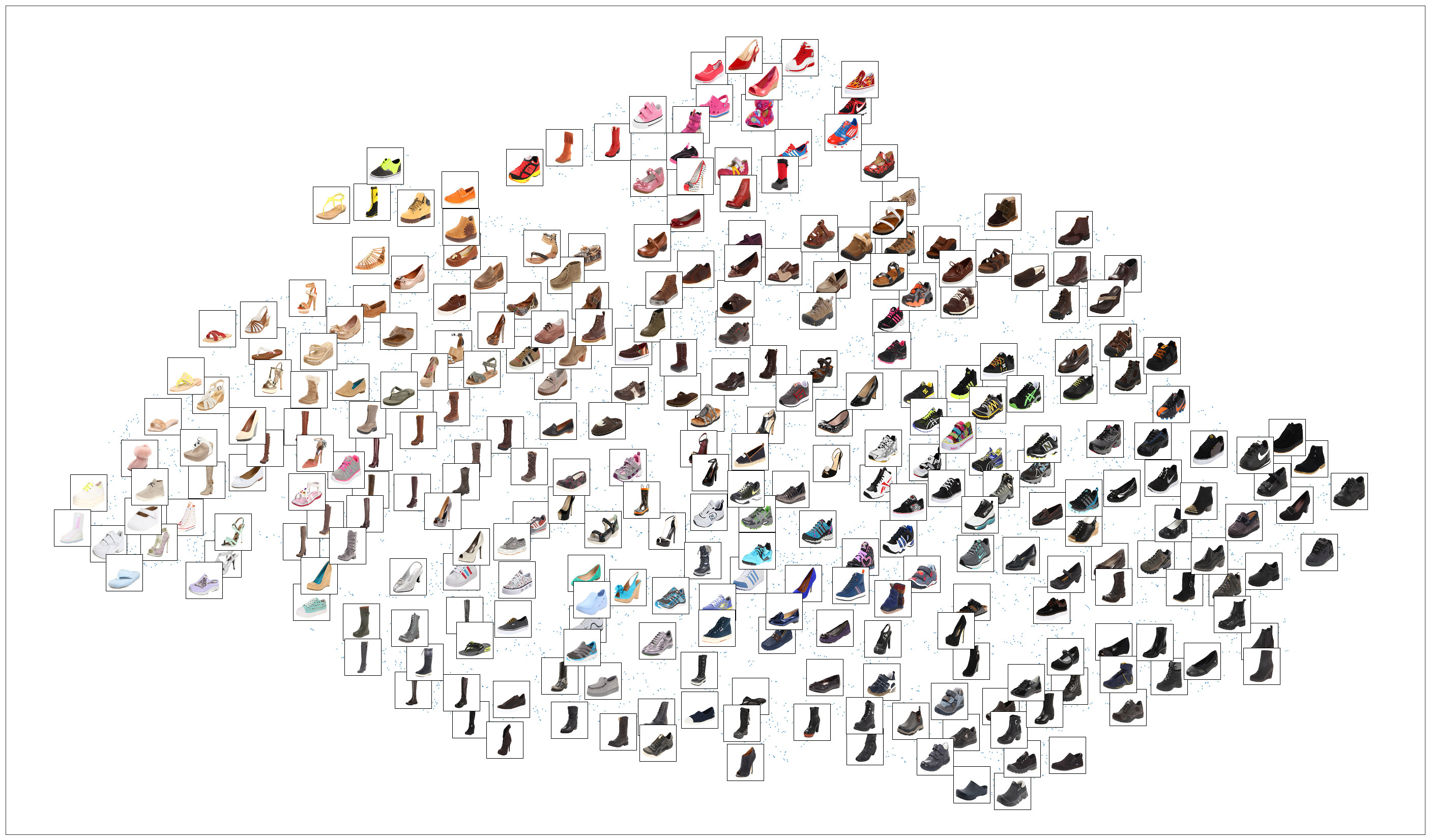}\\
        \caption{Our approach: after finetuning.}
    \end{subfigure}
    \begin{subfigure}[b]{0.63\linewidth}
        \centering
        \includegraphics[width=\linewidth]{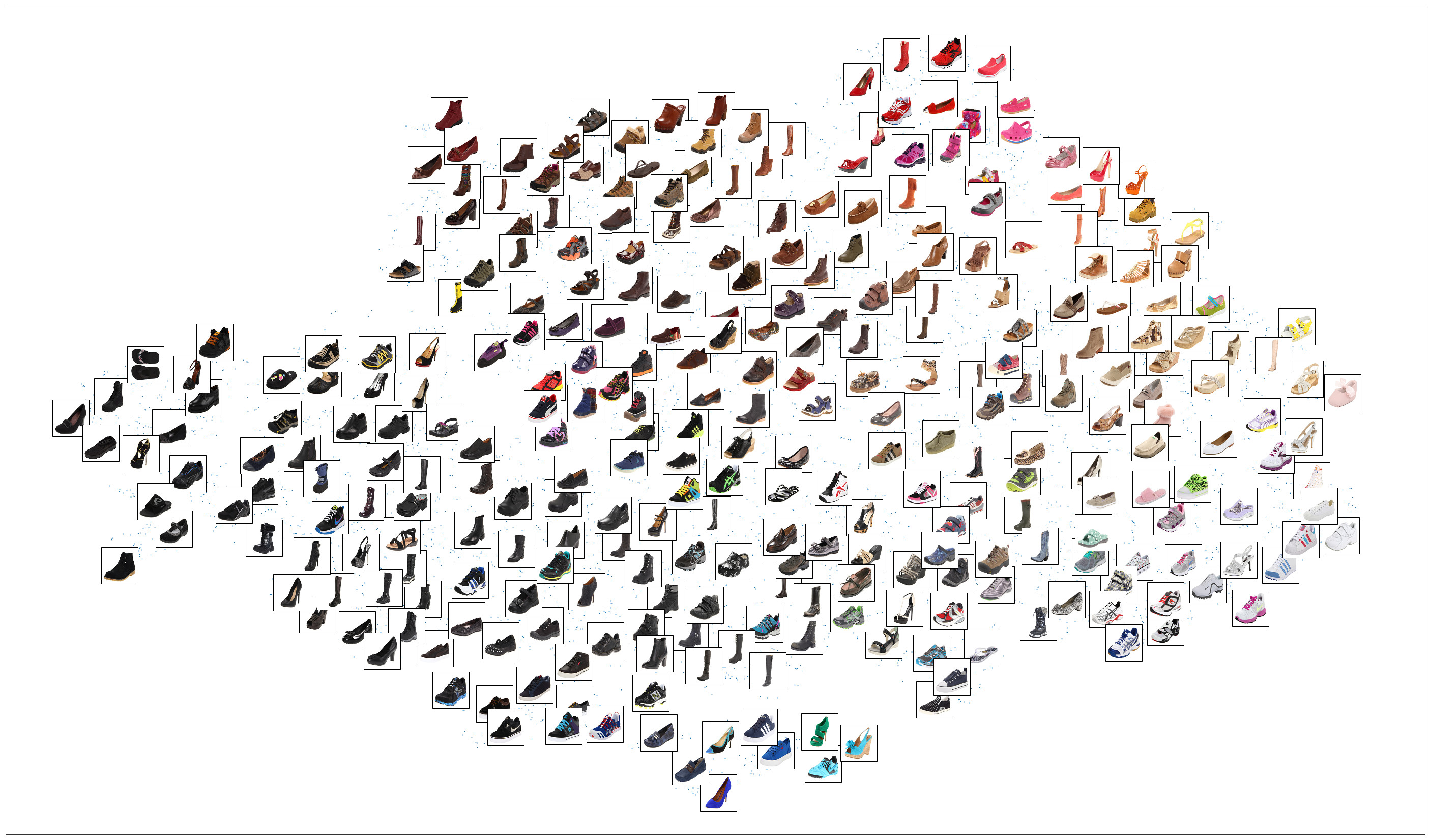}\\
        \caption{BicycleGAN v1 baseline.}
    \end{subfigure}

    \caption{t-SNE plots for the latent style space learned by the style encoder $E$ (a) after style pretraining, (b) after finetuning, and (c) using the BicycleGAN v1 baseline.}
    \label{fig:latent_vis}
\end{figure*}

\begin{figure*}[ht!]
    \centering
    \includegraphics[width=0.85\linewidth]{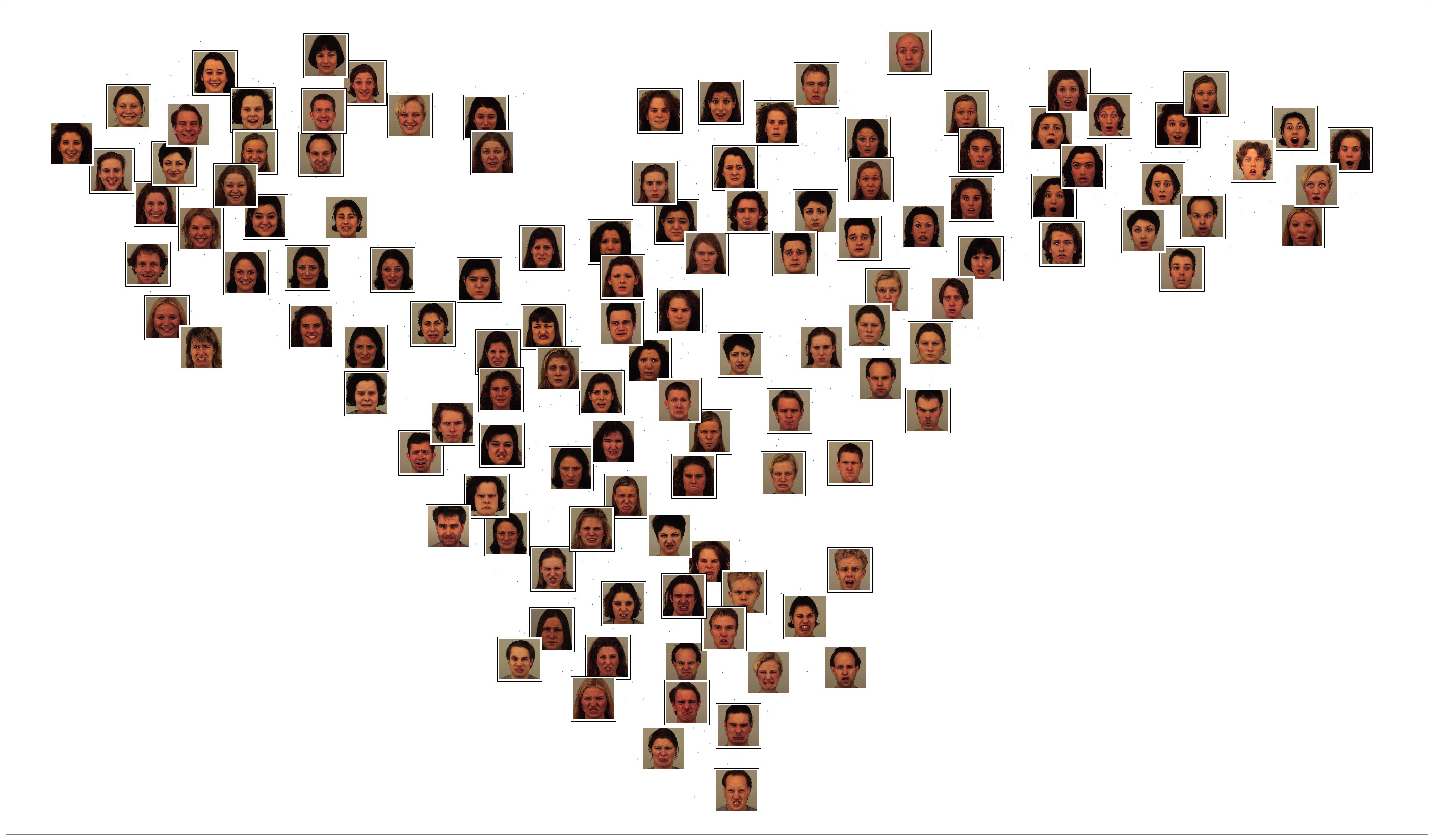}
    \caption{t-SNE plot for the pre-trained latent space learned for facial expressions on a subset of the KDEF dataset.}
    \label{fig:vis_kdef}
\end{figure*}

\subsection{Convergence analysis}
\label{sec:convergence}
Figure~\ref{fig:convergence} compares the convergence of our staged training compared to the BicycleGAN baselines.
The dotted line in the graph marks the transition between stages 2 and 3 of our training (i.e, switching from a fixed pre-trained encoder $E$ to finetuning both $G$ and $E$ together).
We measure the reconstruction error (LPIPS) of the validation set of the edges2handbags dataset as the training progresses.
Results show that with a fixed pre-trained encoder, our staged training starts with higher error than the baselines, but quickly drops to show similar performance as the baselines, and even beats the baselines before switching to stage 3 (marked by a dotted line).
When starting to finetune the encoder $E$, we get a spike in the reconstruction error as the network adapts to the shift in the pre-trained embeddings, but then our staged training steadily widens the performance gap with the baselines.
This shows the importance of the finetuning stage to tweak the pre-trained embeddings to better serve the image synthesis task for the target domain.

\subsection{More quantitative comparison}
We report the Inception Score (IS) computed over the validation set of various datasets in Table~\ref{table:inception}.
Surprisingly, results after finetuning (``ours - stage 3'') are slightly worse than those before finetuning (``ours - stage 2''), but both are still better than the baselines except for the maps dataset. We also note that Inception Score is not very suited to image-to-image translation tasks, since it prefers output diversity with respect to ImageNet classes, not within-class diversity as in our case.

\begin{table}[!ht]
  \caption{Inception score comparison (higher is better) for different datasets.}
  \label{table:inception}
  \centering
  \renewcommand{\arraystretch}{1.2}
  \renewcommand{\tabcolsep}{1.5mm}
  \begin{footnotesize}
    \resizebox{\linewidth}{!}{
      \begin{tabular}{@{}lcccccc@{}}
        \toprule
        \multirow{2}{*}{}  & \multicolumn{1}{c}{handbags} & \multicolumn{1}{c}{shoes} & \multicolumn{1}{c}{facades} & \multicolumn{1}{c}{night2day} & \multicolumn{1}{c}{maps} & \multicolumn{1}{c}{space needle} \\
         \midrule 
        {Bicycle v1}     & 2.13              & 2.83             & 1.41           & 1.65          & 3.26           & 1.82          \\
        {MUNIT-p}     & 2.07              & 2.64             & 1.45           & 1.74          & \textbf{3.57}  & 1.77          \\
        {\textbf{Ours} - stage 2} & \textbf{2.22}     & 2.75    & \textbf{1.61}  & 1.76          & 3.32           & \textbf{1.90}          \\
        {\textbf{Ours} - stage 3} & 2.15              & \textbf{2.85}             & 1.56           & \textbf{1.84} & 3.28           & 1.89          \\

        \bottomrule
      \end{tabular}
    }
  \end{footnotesize}
\end{table}


\subsection{More style interpolations}
\label{sec:more_style_interpolation}
Figure~\ref{fig:app_interpolation_more} shows style interpolation on more datasets.
Notice that, in the edges2handbags results, not only the color is transferred, but also the texture varies from non-smooth to smooth leather. Also, in the maps dataset, the density of bushes varies smoothly.
Figure~\ref{fig:app_interpolation_cmp} further compares our interpolation results with the baselines. Our results show more complex interpolations, as evidenced by the change in lighting and cloud patterns, as well as more faithful style transfer compared to the baselines.


\subsection{Style transfer comparison}
\label{sec:style_transfer_comp}
We compare style transfer performance of our approach against that of the baselines in Figure~\ref{fig:style_transfer_cmp}.
Our approach faithfully captures and transfers colors and weather conditions (including sky and surface lighting) compared to the baselines. 
We attribute the inferior results of the baselines to the reliance on VAEs to train the latent space.
This is because noise added by VAEs means that slight changes to one style would still be mapped to the same point in the latent space, which limits the capacity of low dimensional latent space.
On the other hand, our pre-trained embeddings don't rely on VAEs and hence, can discriminate between more styles.



\subsection{Latent space visualization}
\label{sec:latent_vis}
Figure~\ref{fig:latent_vis_pretrain} visualizes the latent space learned by the style encoder $E$ after pretraining and before finetuning (a), after finetuning (b), and the latent space learned by BicycleGAN~\cite{zhu2017toward} (c).
The embedding learned through pre-training (i.e.~before training the generator $G$) shows meaningful clusters, which verifies the validity of the proposed style-based pre-training.
Finetuning smooths the style clusters and brings the latent space closer to that of BicycleGAN.


\subsection{Encoder pre-training with non-style metrics}
\label{sec:non_style_pretraining}

\begin{figure}[ht!]
    \centering
    \includegraphics[width=0.85\linewidth]{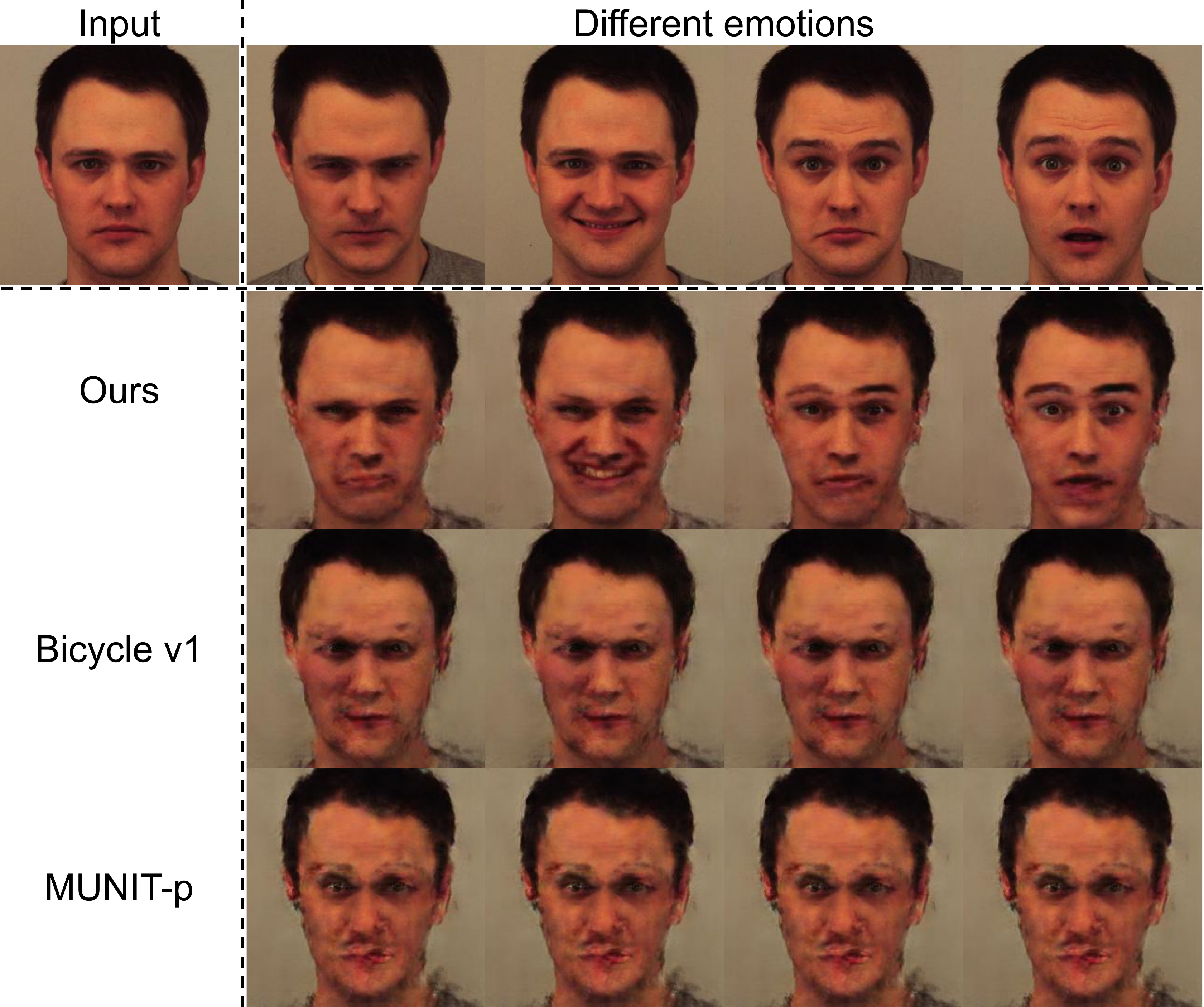}
    \caption{Emotion translation results. First row shows the input image, as well as the ground truth images from which we encode the latent emotion vector for reconstruction.
    Our staged training approach is able to achieve multi-modal synthesis, while the baselines collapse to a single mode.}
    \label{fig:kdef_res}
\end{figure}


\begin{figure*}[t!]
    \centering
    \includegraphics[width=0.8\linewidth]{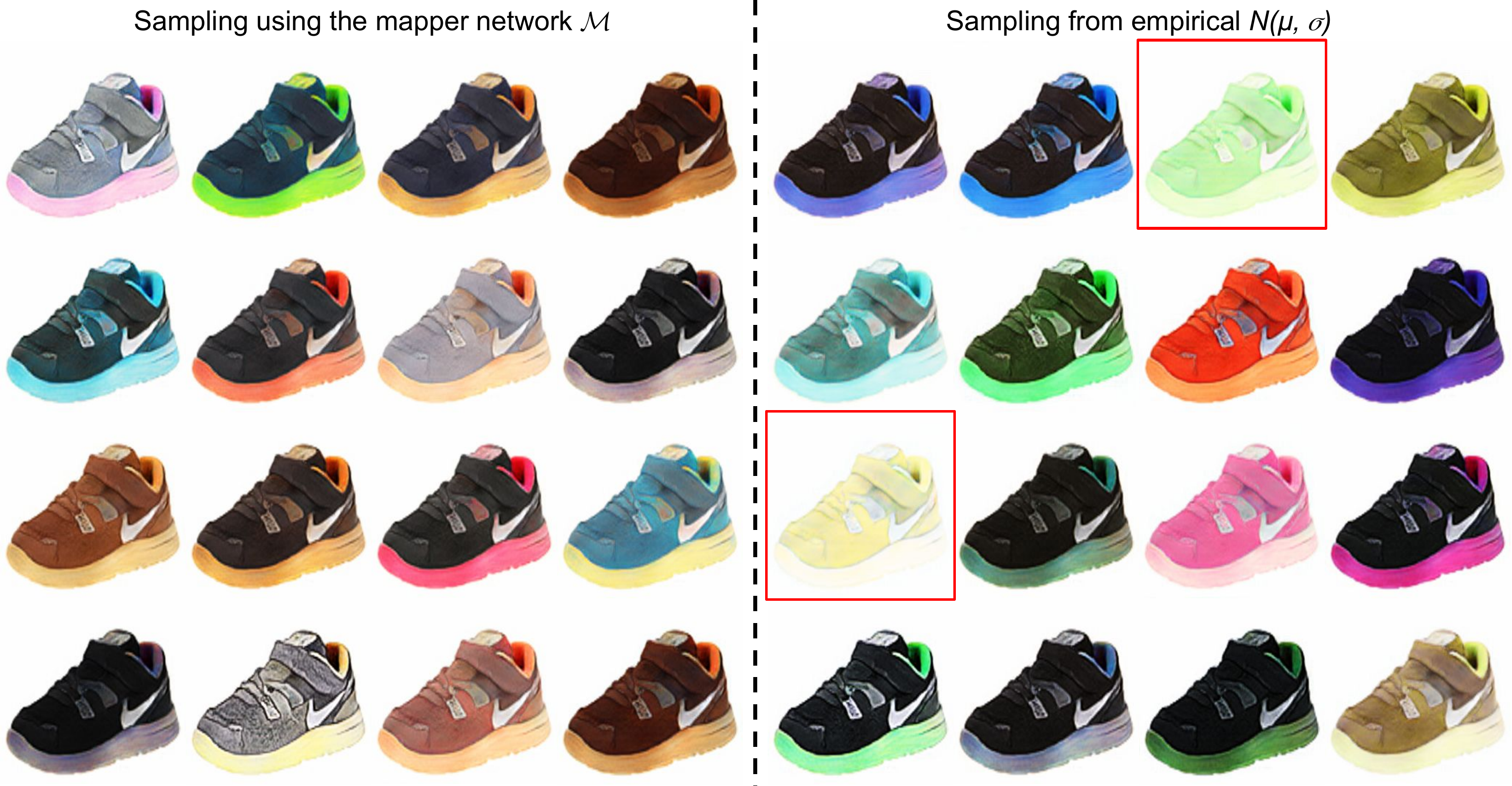}
    \caption{Sixteen randomly sampled styles using both the mapper network $\mathcal{M}$ (left), as well as adhoc sampling from the empirically computed $N(\mu, \sigma)$ distribution of a $L2$-regularized latent space (right). Adhoc sampling could sample bad style codes outside the latent distribution (marked in red).}
    \label{fig:imle_vs_adhoc}
\end{figure*}

Pre-training the encoder using a style-based triplet loss showed to be successful for multi-modal image translation tasks where the variability in the target domain is mainly color-based.
This is shown in the results obtained on several benchmarks, even before the finetuning stage (``ours - stage 2'' in Table 1 of the main text).
We note though that the usage of style-loss as a distance metric for triplet sampling is just one choice and can be replaced with other distance metrics depending on the target application.
Triplet sampling with style distance results in learning an embedding space where images with similar colors/styles lie closely in that space as shown in Section~\ref{sec:latent_vis}.
If, for example, we sample triplets instead based on the distance between VGG-Face~\cite{parkhi2015deep} embeddings, the encoder will learn a latent space which is clustered by identity.
%
%
In this section, we aim to validate that the proposed pre-training strategy can be extended to multi-modal image-to-image translation tasks with non-style variability.
We inspect the task of manipulating facial expressions, where the input is a neutral face, and the output can have other emotions or facial expressions.
For this task, similar emotions should be embedded closely in the latent space.
We therefore use an off-the-shelf facial expression recognition system to compute the emotion similarity/distance between any pair of images.
Specifically, we compute the emotion distance as the euclidean distance between the 512-dimensional feature map of the last layer of a pretrained classification network (e.g.,~\cite{emotion}).
We visualize the learned latent space in Figure~\ref{fig:vis_kdef}, which shows clusters with similar emotions or facial expressions.
We also show example translation results on a holdout set of the front-view images of the KDEF dataset~\cite{kdef} in Figure~\ref{fig:kdef_res}.
We note that the generator successfully learns to manipulate facial expressions based solely on the pre-trained embeddings (without the finetuning stage).
On the other hand, the BicycleGAN-based baselines collapsed to a single mode (over 3 different runs). 
This shows that our staged-training approach is stable and not sensitive to hyper-parameters, unlike the BicycleGAN baselines which will require careful hyper-parameter tuning to work properly on this task.
We also point out that the poor output quality is mainly due to using a pixel-wise reconstruction loss for the generator training, while the input-output pairs in this dataset are not aligned.
We didn't investigate improving the generator training as this is orthogonal to verifying the generalization of encoder pre-training.


\subsection{Style sampling comparison}
\label{sec:sampling_comp}
Figure~\ref{fig:imle_vs_adhoc} compares style sampling using the mapper network $\mathcal{M}$ vs adhoc sampling from the assumed $N(\mu, \sigma)$ of an $L2$-regularized latent space, where $\mu, \sigma$ are empirically computed from the training set. Note that adhoc sampling can sometimes sample bad style codes outside the distribution (e.g.~third image in first row, and first image in third row in the right side of Figure~\ref{fig:imle_vs_adhoc}), since the assumption that a $L2$-regularized space would yield normally distributed latents with zero mean and low standard deviation is not explicitly enforced.

\end{document}